%% file: mtf_cviu.tex
\documentclass[times,twocolumn,final,authoryear]{elsarticle}
\usepackage{ycviu}
\usepackage{framed,multirow}

\usepackage{amssymb}
\usepackage{latexsym}

\usepackage[pagebackref=false, colorlinks=true, citecolor=green]{hyperref} 
\hypersetup{urlcolor=blue}

\usepackage{url}  
\usepackage{times}
\usepackage{epsfig}
\usepackage{epstopdf}
\usepackage{graphicx}
\usepackage{amsmath}

\usepackage{multirow}
\usepackage[table]{xcolor}
\usepackage{tablefootnote}
\usepackage{threeparttable}
\usepackage{listings}
\usepackage{mathtools}
\usepackage{algorithm}
\usepackage{algpseudocode}

\algrenewcommand\textproc{}

\DeclareGraphicsExtensions{.eps}

\definecolor{newcolor}{rgb}{.8,.349,.1}

\newcommand{\abbreviations}[1]{%
	\nonumnote{\textit{Abbreviations:\enspace}#1}}

\journal{Computer Vision and Image Understanding}

\begin{document}

\begin{frontmatter}
	
	\title{Modular Tracking Framework: A unified approach to registration based tracking}
	
	\author[1]{Abhineet \snm{Singh}\corref{cor1}} 
	\cortext[cor1]{Corresponding author: 
		Tel.: +1-587-596-0470;}
	\ead{asingh1@ualberta.ca}
	\author[1]{Martin \snm{Jagersand}}

	\address[1]{Department of Computing Science, University of Alberta, Edmonton T6G2R3, Canada}
	
	\received{1 May 2013}
	\finalform{10 May 2013}
	\accepted{13 May 2013}
	\availableonline{15 May 2013}
	\communicated{S. Sarkar}

	\begin{abstract}

	This paper presents a modular, extensible and highly efficient open source framework for registration based tracking
	called Modular Tracking Framework (MTF).
	Targeted at robotics applications, it is implemented entirely in C++ and designed from the ground up to easily integrate with systems that support any of several major vision and robotics libraries including OpenCV, ROS, ViSP and Eigen.
	It implements more methods, is faster, and more precise than other existing systems.
	Further, the theoretical basis for 
	its design is
	a new way to conceptualize registration based trackers that decomposes them into three constituent sub modules
	- Search Method (SM), Appearance Model (AM) and State Space Model (SSM).
	In the process,
	we integrate many important advances published after Baker \& Matthews' landmark work in 2004.
	
	In addition to being a practical solution for fast and high precision tracking, MTF can also serve as a useful research tool by allowing existing and new methods for any of the sub modules to be studied better.
	When a new method is introduced for one of these, the breakdown can help to experimentally find the combination of methods for the others that is optimum for it.
	By extensive use of generic programming, MTF makes it easy to plug in a new method for any of the sub modules so that it can not only be tested comprehensively with existing methods but also become immediately available for deployment in any project that uses the framework.
	With  16 AMs, 11 SMs and 13 SSMs implemented already, MTF provides over 2000 distinct single layer trackers.
	It also allows two or more of these to be combined together in several ways to create a practically unlimited variety of novel multi layer trackers. 
		\end{abstract}
	
	\begin{keyword}
		\KWD visual tracking\sep image registration\sep robotics\sep software library
		
	\end{keyword}
	
	\abbreviations{
		\textbf{MTF}: Modular Tracking Framework;
		\textbf{SM}: Search Method;
		\textbf{AM}: Appearance Model;
		\textbf{SSM}: State Space Model;
		\textbf{ILM}: ILumination Model;
		\textbf{RBT}: Registration-Based Tracker;
		\textbf{OLT}: Online Learning and detection based Tracker;
		\textbf{LK}: Lucas Kanade;
		\textbf{DOF}: Degrees Of Freedom		
	}
\end{frontmatter}

\rowcolors{2}{gray!25}{white}

\input{introduction}

\input{theoretical_background}
\input{system_design}

\input{examples}

\input{use_cases}

\input{performance}
\input{conclusion}
\input{appendix}

\bibliographystyle{model2-names}
\bibliography{references}

\end{document}

%% file: introduction.tex
\section{Introduction}
\label{introduction}

Fast and high precision visual tracking is crucial to the success of several robotics applications like visual servoing and autonomous navigation. Image patch trackers are usually classified as either registration-based (RBT) or online learning and detection based trackers (OLT). In recent years OLTs have become popular \citep{Wu13benchmark, Kristan2015_vot15} due to their robustness to changes in the object appearance.
This makes OLT suited to long term tracking. However, OLTs are neither precise, nor fast, and commonly estimate only 2 DOF image translation or, adding scale, 3 DOF.
RBTs (Sec. \ref{registration_tracking}) are often more suitable for robotics applications, being several times faster and capable of estimating
higher degrees-of-freedom (DOF) transformations like 6 DOF affine and 8 DOF homography to pixel precision.
Using visual servoing or a calibrated camera, this allows precise full 6 DOF 3D translation and rotation alignments in applications such as robot arm control and augmented reality.

Though several major advances have been made in this domain since the original Lucas Kanade (LK) tracker was introduced almost thirty five years ago \citep{Lucas81lucasKanade}, efficient open source implementations of recent trackers are surprisingly difficult to find. In fact, the only such tracker offered by the popular OpenCV library \citep{opencv_library}, uses a pyramidal implementation of the original 1981 algorithm \citep{Bouguet00_pyr_lk}. Similarly, the more recent ROS library \citep{Quigley09_ros}
currently does not have any package that implements a modern RBT. The XVision system \citep{hager1998_xvision} did introduce a full tracking software framework including a video pipeline. However, it has not been updated for a long time and mainly implements variants of Hager's 1998 algorithm \citep{Hager98parametricModels}.
Even the fairly recent MRPT library \citep{Harris11_robotics_sw_survey} includes only a version of the original LK tracker
apart from a low DOF particle filter based tracker that is too imprecise and slow to be considered relevant for our target applications.
The Visual Servoing Platform (ViSP) library \citep{Marchand05_visp} does provide somewhat similar tracking functionality as the proposed framework but is more limited, slower and not modular, hence harder to extend with new AMs, SMs and SSMs (see Sec. \ref{sec_performance}).

To address the need for a  tracking library targeted at high DOF robotics and augmented reality applications, we introduce Modular Tracking Framework (\textbf{MTF}) \citep{mtfweb}
- a generic system for RBT that provides highly efficient implementations for a large subset of trackers introduced in literature to date and is designed to be easily extensible with additional methods, and compatible with popular pachages such as ROS, OpenCV and ViSP.

MTF conceptualizes an RBT as being composed of three semi independent sub modules - Search Method (\textbf{SM}), Appearance Model (\textbf{AM}) and State Space Model (\textbf{SSM}).
SM is treated here very generally as a way to use the functionality in AM and SSM - through a well defined interface - to solve the tracking problem. 
The three sub modules are semi-independent in the sense that the functional specification of the interface between them is general enough to allow any method implementing one of these to be combined unchanged with many combinations of methods for the other two.

Such an approach can help to address another urgent need in this field - that of unifying the myriad of contributions made in the last three decades so they can be better understood.
When a new RBT is published, it often contributes to only one or two of these sub modules while using existing methods for the rest \citep{singh16_modular_results,singh17_ssim}.
In such cases, the modular decomposition can 
provide a way within which the contributions of the new tracker can be clearly demarcated and thus studied better. 
By following this decomposition closely through extensive use of generic programming, MTF provides a convenient interface to plug in a new method for any sub module and test it against
existing methods for the other two.
This will not only help to compare the new method against existing ones in a more comprehensive way but also make it immediately available to any project that uses MTF.
To facilitate the latter, MTF provides a simple ROS interface for seamless integration with robotics systems along with 
interfaces for Python and MATLAB to aid its use in research applications.

Further, MTF is designed to allow two or more trackers
to be combined in several ways (Sec. \ref{system_design}) to create composite multi-layer trackers that perform better than any of their constituents. 
This approach has been shown to be promising \citep{Zhang2015_rklt,singh17_ssim,singh17_mtf_thesis} in creating trackers that are robust to challenges like illumination changes, fast motion and occlusions by combining the advantages provided by their constituents.
This is also a much easier way to improve tracking performance than designing new algorithms for individual sub modules to handle specific challenges.
An example of such a composite tracker that can be considered as the current state of the art in RBT is the LMES tracker \citep{singh17_mtf_thesis} which has a two layer composite Search Method (SM) created by using LMS (Sec. \ref{stochastic_search}) and ESM (Sec. \ref{gradient_descent}) in cascade (Sec. \ref{sec_composite}) as shown in Fig. \ref{fig_composite_sm_horz}.
This tracker benefits from the larger search radius of stochastic SMs as well as the high precision of gradient based SMs to offer superior performance over either.

To summarize, following are the main contributions of this work:
\begin{itemize}
\item Provide a unifying formulation for RBT that can be seen as an extension of the framework reported in \cite{Baker04lucasKanade_paper} with newer methods.
\item Present a fast tracking library for robotics applications that is based on this formulation and is also easy to extend due to its modular design.
\begin{itemize}
\item
It currently has 16 AMs, 11 SMs and 13 SSMs implemented (Fig. \ref{fig_class_diagram})  and, since each combination of these methods constitutes a distinct tracker, it can be used to run over 2000 single layer trackers.
\item
Any two or more of these trackers can be combined together in a multitude of ways (Sec. \ref{system_design}) to provide virtually unlimited opportunities to create novel multi layer composite trackers.
\end{itemize}
\end{itemize}
This paper extends a preliminary version of this work  \citep{singh17_mtf} by providing detailed function specifications for the main classes in MTF (Sec. \ref{system_design}), examples of existing SMs recast to fit the proposed framework (Sec. \ref{examples}), examples of composite SMs  (Sec. \ref{sec_composite}) and several extensions to the formulation of Baker \& Matthews \citep{Baker04lucasKanade_paper}
(\hyperref[appendix]{appendix}).

The rest of this paper is organized as follows: Section \ref{background} introduces the mathematical basis for the design of MTF while section \ref{system_design} describes the class structure of MTF along with specifications for important functions.
Section \ref{examples} presents several SMs as examples of using the functionality described in section \ref{system_design} to implement the theory of section \ref{background}. Section \ref{use_cases} presents several use cases for MTF while section \ref{sec_performance} provides performance comparison of MTF with another existing library for RBT as well as with state of the art OLTs.
Finally, section \ref{conclusions} concludes with promising avenues for future extensions to this work. The paper includes more technical implementation details than typical in the literature. We feel this is important, as it ultimately affects performance. Someone interested in the broad strokes can read the technical sections cursorily. Someone just wanting to use the implemented trackers can focus on sections \ref{use_cases} and \ref{sec_performance}, while those aiming to do research in tracking systems mainly benefit from the design sections \ref{system_design} and \ref{examples}.

%% file: theoretical_background.tex

\section{Theoretical Background}
\label{background}
\subsection{Notation}
Let $ I_t : \mathbb R^2\mapsto \mathbb R $ refer to an image captured at time $t$. $I_t$ is treated as a smooth function of real values using sub pixel interpolation \citep{Dame10_mi_ict} for non integer locations.
The patch corresponding to the tracked object's location in $ I_t $ is denoted by  $\mathbf{I_t}(\mathbf{x_t}) \in \mathbb R^N $ where $\mathbf{x_t}=[\mathbf{x_{1t}},..., \mathbf{x_{Nt}}]$ with $\mathbf{x_{kt}}=[x_{kt}, y_{kt}]^T \in  \mathbb R^2$ being the Cartesian coordinates of pixel $ k $.

Further, $\mathbf{w}(\mathbf{x}, \mathbf{p_s}) : \mathbb{R}^2 \times \mathbb{R}^S\mapsto \mathbb{R}^2$ denotes a warping function of $ S $ parameters
that represents the set of allowable image motions of the tracked object by specifying the deformations that can be applied to $\mathbf{x_0}$ to align
$\mathbf{I_t}(\mathbf{x_t})=\mathbf{I_t}(\mathbf{w}(\mathbf{x_0},\mathbf{p_{st}}))$
with
$ \mathbf{I_0}(\mathbf{x_0}) $.
Examples of $ \mathbf{w} $ include homography, affine, similitude, isometry and translation \citep{Szeliski2006_fclk_extended}.

Finally $f(\mathbf{I^*}, \mathbf{I^c},\mathbf{p_a} ) : \mathbb{R}^N \times \mathbb{R}^N \times \mathbb{R}^A\mapsto \mathbb{R}$ is a function of $A$ parameters that measures the similarity between two
patches - the reference or template patch $\mathbf{I^*}$
and a candidate patch $ \mathbf{I^c} $.
Examples of $f$ with $A=0$ include sum of squared differences (SSD) \citep{Hager98parametricModels}, sum of conditional variance (SCV) \citep{Richa11_scv_original}, normalized cross correlation (NCC) \citep{Scandaroli2012_ncc_tracking}, mutual information (MI) \citep{Dame10_mi_ict} and cross cumulative residual entropy (CCRE) \citep{Richa12_robust_similarity_measures}.
So far, the only examples with  $A\neq 0$,
to the best of our knowledge,
are those with an illumination model (\textbf{ILM}) \citep{Silveira2007_esm_lighting, Bartoli2008} where $f$ is expressed as $f(\mathbf{I^*}, g(\mathbf{I^c},\mathbf{p_a}))$ with $g : \mathbb{R}^N \times \mathbb{R}^A \mapsto \mathbb{R}^N$ accounting for differences in lighting conditions under which $I_0$ and $I_t$ were captured.
\subsection{Registration based tracking}
\label{registration_tracking}

\begin{figure}[t] 
	\centering
	\includegraphics[width=0.49\textwidth]{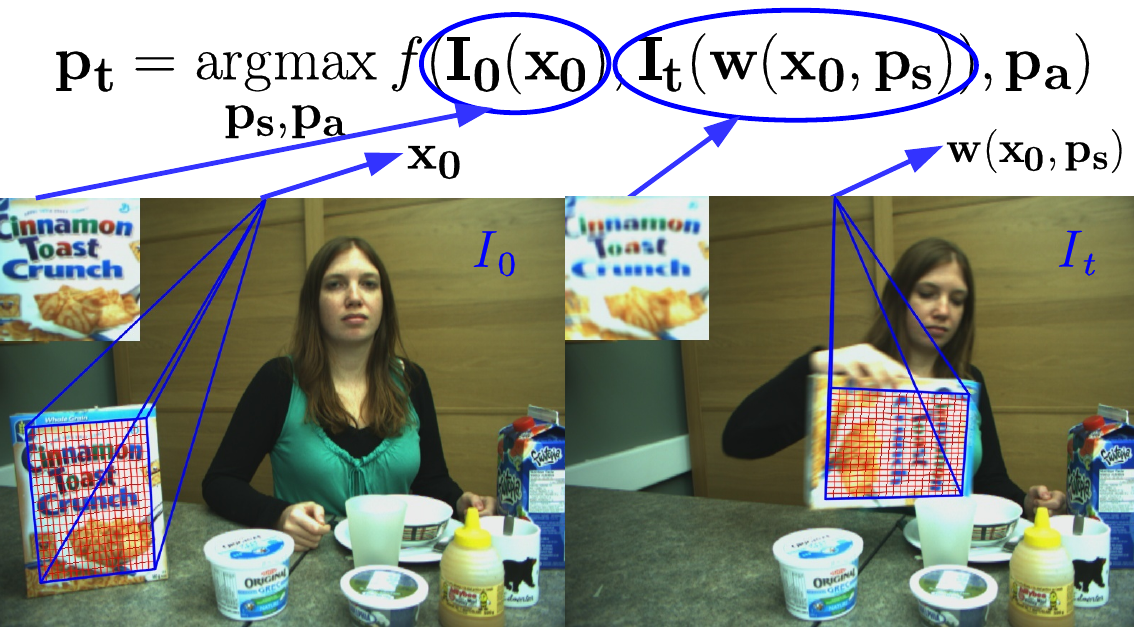}
	\caption{Two frames from a sequence showing the different components of RBT. The grid of points $ \mathbf{x} $ where pixel values are extracted are shown in red. The function $\bf w$ warps the red grid on $\bf I_t$ to match (align with) the template $\bf I_0$ as shown in the top left corners. 
		For better visibility, the grid is sampled at $ 25\times 25 $ though higher resolutions may be used in practice.}
	\label{fig_rbt_demo}
\end{figure}

RBT aligns an image template $ I_0 $ with the corresponding pixels in the current image $ I_t $, see Fig \ref{fig_rbt_demo}. Mathematically, the goal is to find a warp function $ \mathbf{w}(\mathbf{p}) $ that transform the pixel coordinates to match the image patches.
If perfect intensity constancy was possible then $ I_0 = I_t(\mathbf{w}(\mathbf{p})) $.
In real world videos, the appearance of the template will vary, so equality is not possible, and one of several possible Appearance Modules - AM, denoted with $ f $ above, are used for the similarity metric.

Using the notation in the last section, RBT can be formulated (Eq. \ref{tracking}) as a search problem where the goal is to find the optimal parameters $\mathbf{p_t}=[\mathbf{p_{st}}, \mathbf{p_{at}}]\in\mathbb{R}^{S+A}$
that maximize the similarity, measured by $f$, between the target patch 
$\mathbf{I^*} = \mathbf{I_0}(\mathbf{x_0})$
and the warped image patch $\mathbf{I^c}=\mathbf{I_t}(\mathbf{w}(\mathbf{x_0},\mathbf{p_t}))$, that is,
\begin{equation}
\begin{aligned}
\label{tracking}
\mathbf{p_t} = \underset{\mathbf{p_s},\mathbf{p_a} } {\mathrm{argmax}} ~f(\mathbf{I_0}(\mathbf{x_0}),\mathbf{I_t}(\mathbf{w}(\mathbf{x_0},\mathbf{p_s})), \mathbf{p_a})
\end{aligned}
\end{equation}

As has been observed before \citep{Szeliski2006_fclk_extended, Richa12_robust_similarity_measures}, this formulation gives rise to an intuitive way to decompose the tracking task into three modules - the similarity metric $ f $, the warping function $ \mathbf{w} $ and  the optimization approach. We refer to these respectively as Appearance Model- AM, State-Space Model - SSM and Search Method - SM.
These can be designed to be  semi independent in the sense that any given optimizer can be applied unchanged to several combinations of methods for the other two modules which in turn interact only through a well defined and consistent interface. 

\begin{figure}[t]
	\centering
	\includegraphics[width=0.49\textwidth]{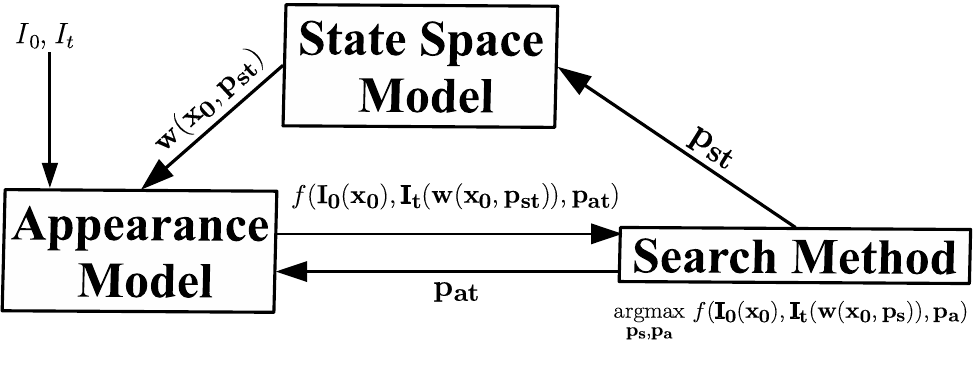}
	\caption{Decomposition of RBT showing the interaction between the resultant sub modules}
	\label{fig_rbt_submodules} 
\end{figure}

A pictorial representation of the meanings of various components in this equation is shown in Fig. \ref{fig_rbt_demo}.
As has been observed before \citep{Szeliski2006_fclk_extended, Richa12_robust_similarity_measures}, this formulation gives rise to an intuitive way to decompose the tracking task into three modules - the similarity metric $ f $, the warping function $ \mathbf{w} $ and  the optimization approach. These can be designed to be  semi independent in the sense that any given optimizer can be applied unchanged to several combinations of methods for the other two modules which in turn interact only through a well defined and consistent interface. 

Fig. \ref{fig_rbt_submodules} shows the effective flow of information between the three sub modules though in practice SM serves as the interface between AM and SSM, which do not interact directly.
Following steps are performed for each frame $ I_t $:
\begin{enumerate}
\item SM computes the optimum parameters for $ \mathbf{w} $ and $ f $ and passes these respectively to the SSM and AM.
\item SSM then warps the initial grid points $ \mathbf{x_0} $ using these parameters and passes the resultant points $ \mathbf{w}(\mathbf{x_0}, \mathbf{p_{st}}) $ to the AM 
\item AM extracts pixel values at these points and computes the similarity of the resultant patch with the template using $ \mathbf{p_{at}} $ which it then passes back to the SM.
\item SM uses this similarity to find the parameters $\mathbf{ p_{t+1}} $ that maximize it for the next frame $ I_{t+1} $.
\end{enumerate}

\subsection{Gradient Descent and the Chain Rule}
\label{gradient_descent}
Though several types of SMs
have been reported
in literature, gradient descent based methods \citep{Lucas81lucasKanade} are most widely used due to their speed and simplicity.
As mentioned in \citep{Baker04lucasKanade_paper},
the LK tracker can be formulated in four different ways depending on which image is searched for the warped patch - $ I_0 $ or $ I_t $ - and how $\mathbf{p_s}$ is updated in each iteration - additive or compositional. The four resultant formulations are thus called Forward Additive (\textbf{FALK}) \citep{Lucas81lucasKanade}, Inverse Additive (\textbf{IALK}) \citep{Hager98parametricModels}, Forward Compositional (\textbf{FCLK}) \citep{Shum00_fc} and Inverse Compositional (\textbf{ICLK}) \citep{Baker01ict}. There is also a more recent approach called Efficient Second order Minimization (\textbf{ESM}) \citep{Benhimane07_esm_journal} that tries to make the best of both ICLK and FCLK by using information from both $ I_0 $ and $ I_t $.

What all these methods have in common is that they solve Eq \ref{tracking} by estimating an incremental update $ \Delta\mathbf{p_t} $ to the optimal parameters $  \mathbf{p_{t-1}} $
at time $ t-1 $
using some variant of the Newton method as:
\begin{equation}
\label{eq_nwt_method}
\Delta\mathbf{p_t}=-\hat{\mathbf{H}}^{-1}\hat{\mathbf{J}}^T
\end{equation}
where  $ \hat{\mathbf{J}} $ and $ \hat{\mathbf{H}} $ respectively are estimates for the
Jacobian  $\mathbf{J} = \partial f/\partial \mathbf{p}$
and the
Hessian $\mathbf{H} = \partial^2 f/\partial \mathbf{p}^2$ of $ f $ w.r.t. $ \mathbf{p} $. 
For any formulation that seeks to decompose this class of trackers (among others)
in the aforementioned manner,
the chain rule for first and second order derivatives is indispensable and
the resultant decompositions for $  \mathbf{J} $ and $  \mathbf{H} $ are given by Eqs. \ref{eq_jac_basic} and \ref{eq_hess_basic} respectively, assuming $ A=0 $ (or $  \mathbf{p}= \mathbf{p_s} $) for simplicity.
\begin{align}
\label{eq_jac_basic}
 \mathbf{J} = 
 \dfrac{\partial f(
 \mathbf{I}(
 \mathbf{w}(\mathbf{p}) 
 )
 )}
 {\partial \mathbf{p}} = 
\dfrac{\partial f}{\partial \mathbf{I}} 
\nabla \mathbf{I}
\dfrac{\partial  \mathbf{w}}{\partial \mathbf{p}}
\end{align}

\begin{align}
\label{eq_hess_basic}
\mathbf{H}
=
\dfrac{\partial \mathbf{I}}{\partial \mathbf{p}}^T
\dfrac{\partial^2 f}{\partial \mathbf{I}^2}
\dfrac{\partial \mathbf{I}}{\partial \mathbf{p}}
+
\dfrac{\partial f}{\partial \mathbf{I}}
\dfrac{\partial^2 \mathbf{I}}{\partial \mathbf{p}^2}
\end{align} 
with
$
\dfrac{\partial \mathbf{I}}{\partial \mathbf{p}}
=
\nabla \mathbf{I}
\dfrac{\partial\mathbf{w}}{\partial \mathbf{p}}
$
and
$
\dfrac{\partial^2 \mathbf{I}}{\partial \mathbf{p}^2}
=
\dfrac{\partial\mathbf{w}}{\partial \mathbf{p}}^T
\nabla^2\mathbf{I}
\dfrac{\partial \mathbf{w}}{\partial \mathbf{p}}
+
\nabla\mathbf{I}
\dfrac{\partial^2 \mathbf{w}}{\partial \mathbf{p}^2}
$. 
It follows that
the AM computes terms involving $  \mathbf{I} $ and $ f $ ($ \nabla \mathbf{I} $, $ \nabla^2\mathbf{I} $, $ \partial f/\partial \mathbf{I} $ and $ \partial^2 f/\partial \mathbf{I}^2 $ )
while
the SSM computes those with $  \mathbf{w} $ ($ \partial\mathbf{w}/\partial \mathbf{p}$, $ \partial^2 \mathbf{w}/\partial \mathbf{p}^2 $).
Further, these generic expressions do not give the whole scope of the decompositions since
the exact forms of $ \hat{\mathbf{J}} $ and $ \hat{\mathbf{H}} $ as well as the way these are split vary for different variants of LK.
The reader is referred to \citep{Baker04lucasKanade_paper} for more details though formulations relevant to the functions in MTF (Tables \ref{tab_am_func_specification} and \ref{tab_ssm_func_specification}), including several extensions to \citep{Baker04lucasKanade_paper}, are also presented in the \hyperref[appendix]{appendix}.
\subsection{Stochastic Search}
\label{stochastic_search}
A limitation of gradient descent type SMs
is that they are prone to getting stuck in local maxima of $ f $ especially when the object's appearance changes due to factors like occlusions, motion blur or illumination variations. An alternative approach to avoid this problem is to use stochastic search so as to cover a larger portion of the search space of $ \mathbf{p} $. There are currently four main SMs in this category in MTF - 
particle filter (\textbf{PF}) \citep{Kwon2014_sl3_aff_pf}, nearest neighbor (\textbf{NN}) \citep{Dick13nn}, least median of squares (\textbf{LMS}) \citep{Rousseeuw84_lmeds}  and random sample consensus (\textbf{RANSAC}) \citep{Zhang2015_rklt}.

These SMs work by generating a set of random samples for $ \mathbf{p} $ and evaluating the goodness of each by some measure of similarity with the template.
NN and PF generate samples directly by drawing them from an appropriate Gaussian distribution while LMS and RANSAC obtain them indirectly by finding the best fit parameters that explain the transformation between two sets of corresponding points.
The performance of these SMs thus thus depends mostly on the number and quality of stochastic samples used. While the former is limited only by the available computational resources, the latter is a bit harder to guarantee for a general SSM/AM. For methods that draw samples from a Gaussian distribution, the quality thereof is determined by the covariance matrix used and, to the best of our knowledge, no widely accepted method exists to estimate it in the general case. Most works either use heuristics or perform extensive hand tuning to get acceptable results, sometimes even using different values for each tested sequence \citep{Kwon2014_sl3_aff_pf}. 

Given this, a reasonable way to decompose these methods to fit our framework is to delegate the responsibility of generating the set of samples and estimating its mean entirely to the SSM
and AM
while letting the latter evaluate the suitability of each sample by providing the likelihood of the corresponding patch.
Such a decomposition ensures both theoretical validity and good performance in practice
since the definition of what constitutes a good sample and how the mean of a sample set is to be evaluated depends on the SSM/AM, as do any heuristics for generating these samples
(like the variance for each component of $ \mathbf{p} $).


%% file: system_design.tex
\section{System Design}
\label{system_design}
\begin{figure}
\begin{center}
\setlength{\belowcaptionskip}{-10pt}
\includegraphics[width=0.48\textwidth]{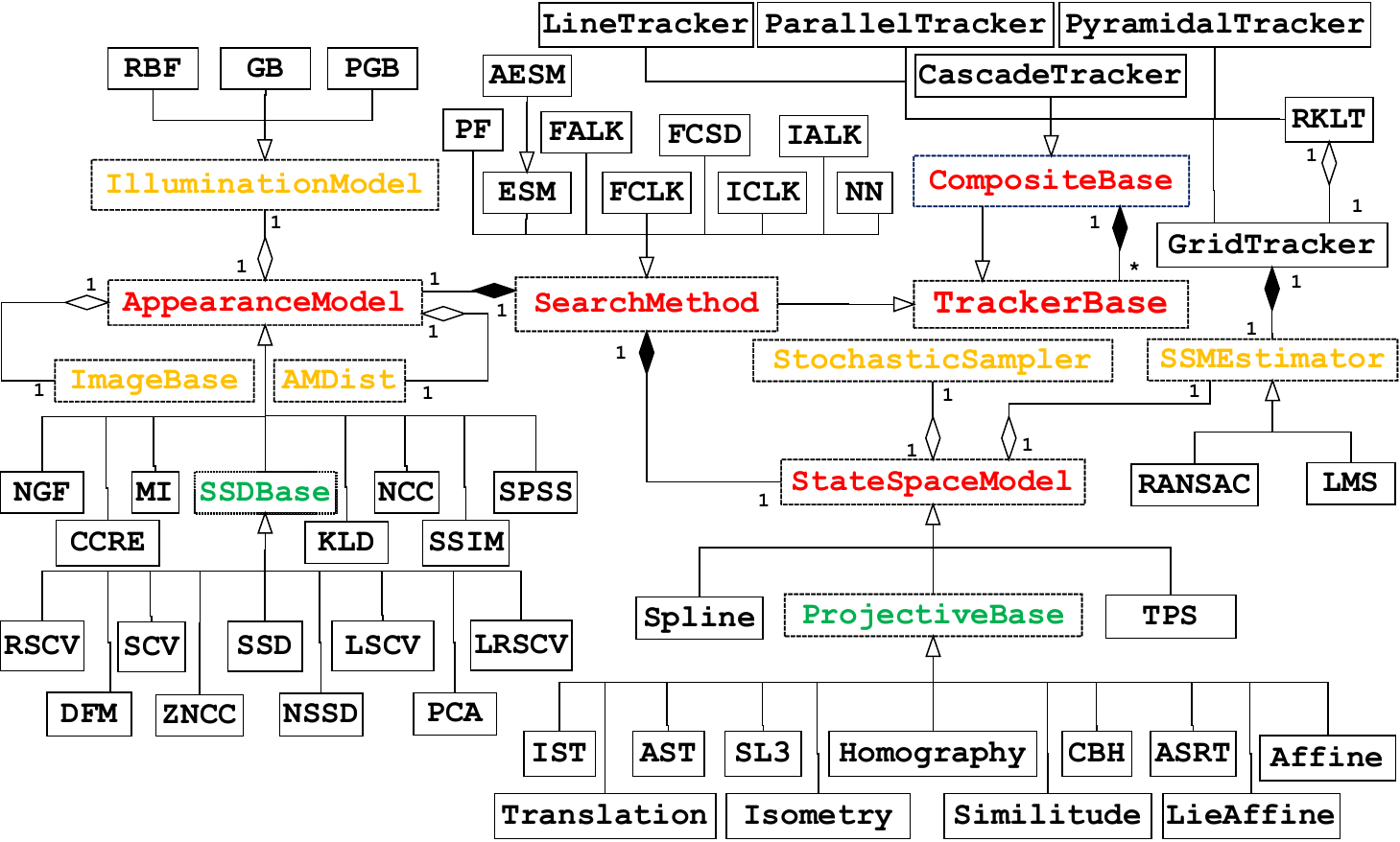}\\
\caption{MTF Class Diagram showing all models currently implemented. Pure and partially abstract classes are respectively shown in red and green while concrete classes are in black. 
Classes that are sub parts of \texttt{AM} and \texttt{SSM} are in yellow. The meanings of acronyms not defined in text can be found in \citep{singh17_mtf_thesis, mtfweb}.
}
\label{fig_class_diagram}
\end{center}
\end{figure}
\subsection{Overview}
\label{sec_sys_design_overview}

As shown in the class diagram in Fig. \ref{fig_class_diagram}, MTF closely follows the decomposition described in the previous section and has three abstract base classes corresponding to the three sub modules - \texttt{SearchMethod}, \texttt{AppearanceModel} and \texttt{StateSpaceModel}. \footnote{For brevity, these will be referred to as \texttt{SM}, \texttt{AM} and \texttt{SSM} respectively with the font serving to distinguish the \textit{classes} from the corresponding \textit{concepts}.} 
Of these, only  \texttt{SM} is a generic/templated class that is templated on specializations of the other two classes. A concrete tracker, defined as a particular combination of the three sub modules, thus corresponds to a subclass of \texttt{SM} that has been instantiated with subclasses of \texttt{AM} and \texttt{SSM}. 

A particular SM in this formulation is defined only by its objective - to find the $ \mathbf{p} $ that maximizes the similarity measure defined by the AM. Thus, different implementations of \texttt{SM} can cover a potentially wide range of methods that have little in common. As a result, \texttt{SM} is the least specific of these classes and only provides functions to initialize, update and reset the tracker along with accessors to obtain its current state. In fact, an SM is regarded in this framework simply as one way to \textit{use} the methods provided by the other two sub modules in order to accomplish the above objective. The idea is to abstract out as much computation from the SM to the AM/SSM as possible so as to make for a general purpose tracker. 
Therefore, this section describes only \texttt{AM} and \texttt{SSM} in detail while some of the SMs currently available in MTF are presented in the next section as examples of using the functionality described here to carry out the search in different ways. 

Another consequence of this conceptual impreciseness of \texttt{SM} is that a specific SM may use only a small subset of the functionality provided by \texttt{AM}/\texttt{SSM}. 
For instance, gradient descent type SMs do not use the random sampling functions of \texttt{SSM} and conversely, stochastic SMs do not use the derivative functions required by the former.
This has two further implications. Firstly, the functionality set out in \texttt{AM} and \texttt{SSM} is not fixed but can change depending on the requirements of an SM, i.e. if a new SM is to be implemented that requires some functionality not present in the current specifications, the respective class can be extended to support it - as long as such an extension makes logical sense within the definition of that class. Secondly, it is not necessary for all combinations of AMs and SSMs to support all SMs. For instance a similarity measure does not need to be differentiable to be a valid AM as long as it is understood that it cannot be used with SMs that require derivatives.

In the broadest sense, the division of functionality between \texttt{AM} and \texttt{SSM} described next can be seen as \texttt{AM} being responsible for everything to do with the image $ I $, the sampled patch $ \mathbf{I}(\mathbf{x}) $ and the similarity $ f $ computed using it, while \texttt{SSM} handles the actual \textit{points} $ \mathbf{x} $ at which the patch is sampled along with the warping function $ \mathbf{w} $ that defines $ \mathbf{x} $ in terms of $ \mathbf{x_0} $ and $ \mathbf{p_s} $.
\subsection{Composite Tracking}
\label{sec_composite}

\begin{figure}
	\begin{center}
		\includegraphics[width=0.49\textwidth]{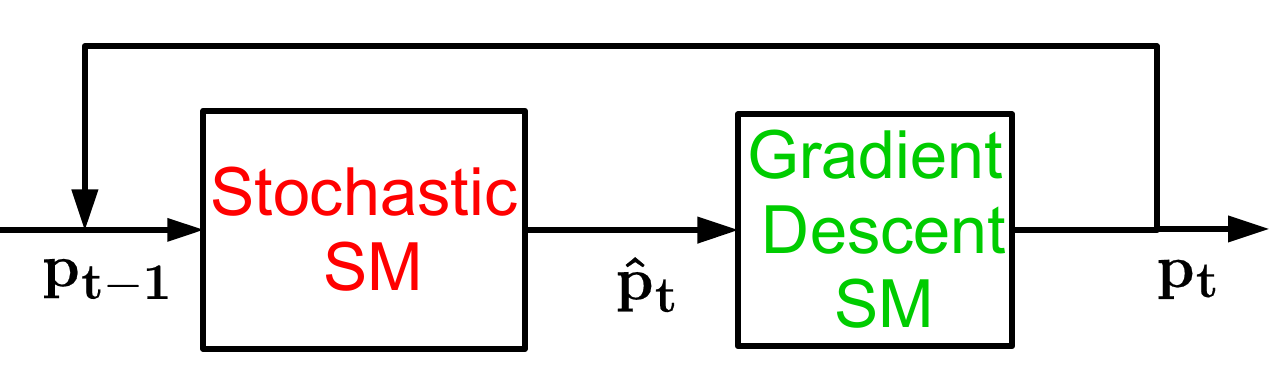}\\
		\caption{A stochastic and gradient descent SM in cascade. The rough estimate produced by the former is refined by the latter and then fed back to the former to be used as the starting search point for the next frame.}
		\label{fig_composite_sm_horz}
	\end{center}
\end{figure}

It may be noted that \texttt{SM} itself derives from a non generic base class called \texttt{TrackerBase} for convenient creation and interfacing of objects corresponding to heterogeneous trackers, including those external to MTF
\footnote{Several state of the art OLTs with publicly available C++ code - DSST \citep{Danelljan2016_dsst}, KCF \citep{henriques2015high_kcf}, CMT \citep{Nebehay2015_cmt}, Struck \citep{Hare16_struck}, TLD  \citep{Kalal12tld}, MIL \citep{Babenko2011_mil}, RCT  \citep{Zhang2012_rct}, FragTrack \citep{Adam2006_fragtrack}, GOTURN \citep{Held2016_goturn} - have already been integrated with MTF.}
,
so that they can be run simultaneously and their results combined to create composite trackers that are more robust than any of their components.
Allowing a diverse variety of trackers to integrate seamlessly is one of the core design objectives of MTF
and this is emphasized by having such composite trackers derive from a separate base class called \texttt{CompositeBase} which in turn derives from \texttt{TrackerBase} while also containing several instances of it. 
Individual RBTs are well known to be prone to failures and since more than three decades of research has failed to make significant improvements in this regard \citep{singh17_ssim}, this approach seems to be one of the more promising ones  \citep{Zhang2015_rklt, singh17_ssim}.
MTF has thus been designed to facilitate work in this direction.

Five composite trackers have currently been implemented:
\begin{itemize}
\item \texttt{GridTracker}:
This corresponds to the stochastic SMs based on indirect sampling mentioned in Sec. \ref{stochastic_search}.
This uses a grid of (typically low DOF) trackers such that each tracks a different sub patch within the tracked object. 
The independent results of these trackers are then combined by a robust estimator provided by the SSM to estimate the best fit warp that gives the overall location of the patch.

\item \texttt{LineTracker}: This identifies straight lines in the object of interest and uses 2 DOF trackers to track multiple points on each line. The outputs of these trackers are then used to estimate the best fit line assuming linearity of points as an invariant property under the warp that the object patch has undergone.
Constraints between different lines, such as parallelism, can also be enforced to improve this estimation further.
By resetting the trackers to their expected positions on these lines, any drift can be compensated for.

\item \texttt{ParallelTracker}: This one runs multiple trackers in parallel to track the same patch and then combines their outputs to produce a more robust estimate of the location of the tracked object. Many methods may be used to combine the locations produced by the different trackers, with a simple example being to take the mean of the bounding box corners. 

\item \texttt{PyramidalTracker}: This one builds a Gaussian image pyramid \citep{Bouguet00_pyr_lk} and then tracks each level of the pyramid through a different instance of (usually) the same tracker. The output of the tracker at level $ n $, after appropriate scaling, is used as starting point for the one at level $ n + 1 $ .

\item \texttt{CascadeTracker}: This tracks the same patch using multiple trackers too, but here the output of the tracker at each layer of the cascade is used as the starting point for the tracker at the next layer. The output of the last layer is fed back to the first one to make a closed loop system.
Fig. \ref{fig_composite_sm_horz} shows an example of this arrangement where two SMs are used in cascade. 
The top-performing LMES is an example (Sec. \ref{sec_performance}) of this. 
\end{itemize}

Though not shown in Fig. \ref{fig_class_diagram}, several of these composite trackers also have specialized variants where all the constituent trackers have the same AM and SSM and differ only in their SM. Since all SMs derive from \texttt{SM}, this specialization makes it possible for the constituent trackers to utilize the functionality available in \texttt{SM} but not in \texttt{TrackerBase} in the ways that they interact with and benefit from each other.  \texttt{TrackerBase} is designed to allow third party trackers including OLTs to integrate with MTF and so does not include functions specific to RBTs.
An example of such a functionality is selective pixel integration \citep{Dellaert1999} which is used by the RKLT tracker \citep{Zhang2015_rklt} to filter out parts of the patch that do not agree with the template due to partial occlusions or localized illumination changes.

It is emphasized here that any two or more of the composite trackers can themselves be combined together in arbitrary ways. For example, it is straightforward to create a cascade tracker where each layer is itself a pyramidal tracker. It would be equally easy to create a pyramidal tracker where each level is tracked by a cascade tracker. Different levels of the pyramid might even be tracked by different cascade arrangements just as different layers of the cascade in the previous example might differ in the number of pyramidal levels within the corresponding trackers.


\subsection{\em{\texttt{AppearanceModel}}}
\label{sec_am}

\begin{figure}
	\begin{center}
		\includegraphics[width=0.5\textwidth]{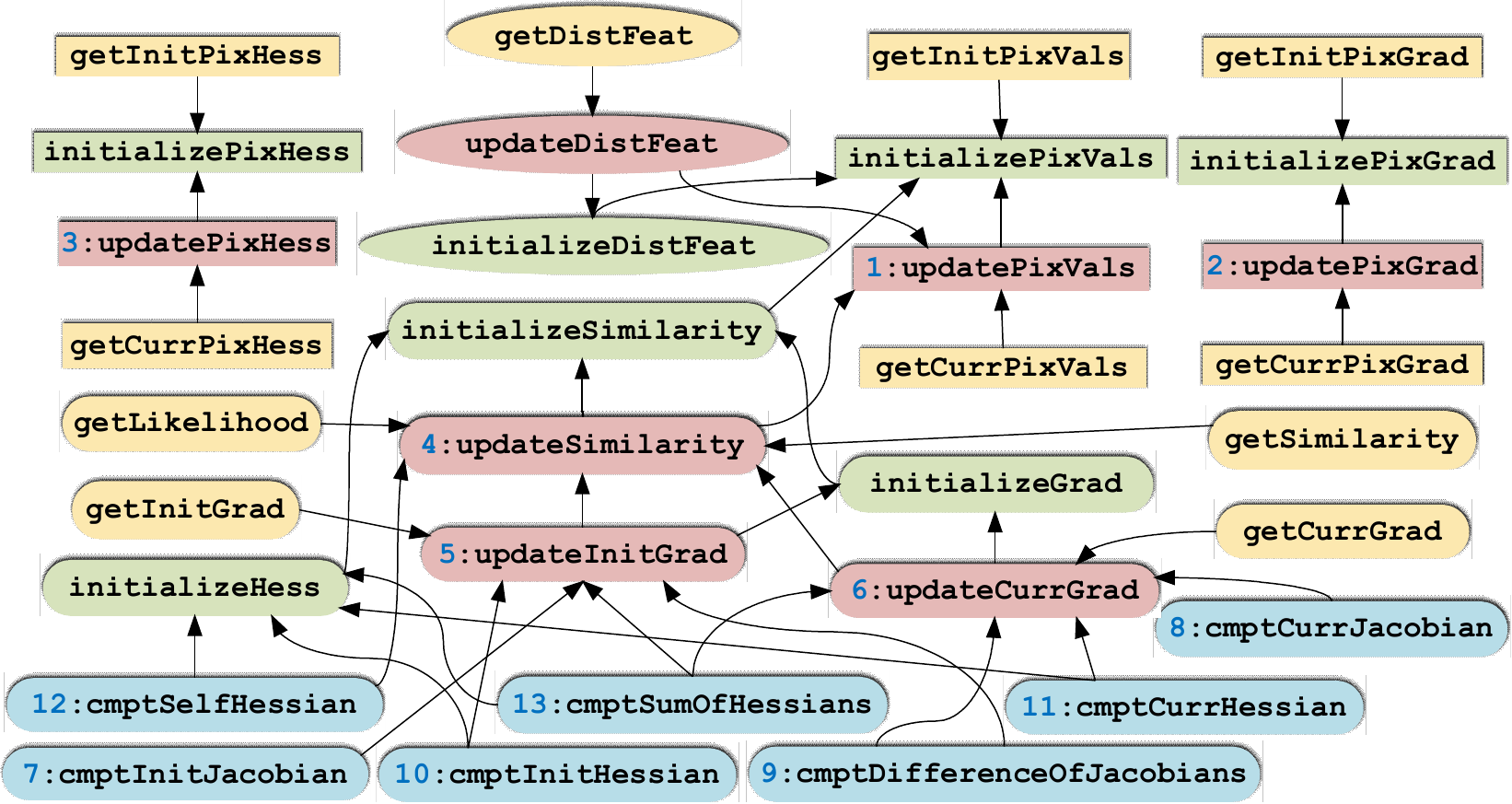}
		\caption{Dependency relationships between various functions in \texttt{AM}: an arrow pointing from A to B means that A depends on B. Color of a function box denotes its type - \textcolor{green}{green}: initializing; \textcolor{red}{red}: updating; \textcolor{blue}{blue}: interfacing and \textcolor{yellow}{yellow}: accessor function. Shape of a function box represents the part of \texttt{AM} it belongs to - rectangle: Image Operations; rounded rectangle: Similarity Functions; ellipse: Distance Feature. The numbers attached to some of the nodes refer Table \ref{tab_am_func_specification}.}
		\label{fig_am_dependency}
	\end{center}
\end{figure}

This class can be divided into three main parts with each defined as a set of variables dependent on $ I_0 $ and $ I_t $ with a corresponding \texttt{\small{initialize}} and \texttt{\small{update}} function for each. The division is mainly conceptual and methods in different parts are free to interact with each other in practice. Table \ref{tab_am_func_specification} presents a brief specification of some important methods in \texttt{AM}.
\begin{table}[t]
\scriptsize
\centering
\begin{threeparttable}
    \caption{Specifications for important methods in \texttt{AM}. IDs in first column refer Fig. \ref{fig_am_dependency}}
    \label{tab_am_func_specification}
    \begin{tabular}{|l|l|l|}	
     \hline
    \rowcolor{gray!50}
    ID & Inputs & Output/Variable updated \\    
    \color{blue}1 & $ \mathbf{x_t}$      & $\mathbf{I_t}(\mathbf{x_t})  $
    \\
    \color{blue}2 &  $ \mathbf{x_t}$      &  $ \nabla\mathbf{I_t} $
    \\
    \color{blue}3 & $ \mathbf{x_t}$      &  $ \nabla^2\mathbf{I_t} $
    \\
    \color{blue}4 & None      & $ f(\mathbf{I_0}, \mathbf{I_t}) $
    \\
    \color{blue}5 & None      &  $ \dfrac{\partial f(\mathbf{I_0}, \mathbf{I_t})}{\partial \mathbf{I_0}} $ 
    \\[2ex]
    \color{blue}6 & None      & $ \dfrac{\partial f(\mathbf{I_0}, \mathbf{I_t})}{\partial \mathbf{I_t}} $
    \\[2ex]
    \color{blue}7 &  $ \dfrac{\partial \mathbf{I_0}}{\partial \mathbf{p_s}} $      & $ \dfrac{\partial f(\mathbf{I_0}(\mathbf{p}), \mathbf{I_t})}{\partial \mathbf{p}} $  (Eq.  \ref{eq_jac_iclk})
    \\[2ex]
    \color{blue}8 &  $ \dfrac{\partial \mathbf{I_t}}{\partial \mathbf{p_s}} $      & $ \dfrac{\partial f(\mathbf{I_0}, \mathbf{I_t}(\mathbf{p}))}{\partial \mathbf{p}} $ (Eq.  \ref{eq_jac_falk}, \ref{eq_jac_fclk})
    \\[2ex]
    \color{blue}9 & $ \dfrac{\partial \mathbf{I_0}}{\partial \mathbf{p_s}} $, $ \dfrac{\partial \mathbf{I_t}}{\partial \mathbf{p_s}} $      & $  \dfrac{\partial f(\mathbf{I_0}, \mathbf{I_t}(\mathbf{p}))}{\partial \mathbf{p}} - \dfrac{\partial f(\mathbf{I_0}(\mathbf{p}), \mathbf{I_t})}{\partial \mathbf{p}} $  (Eq.  \ref{eq_jac_esm})
    \\[2ex]
    \color{blue}10\tnote{1} & $ \dfrac{\partial \mathbf{I_0}}{\partial \mathbf{p_s}} $,$ \dfrac{\partial^2 \mathbf{I_0}}{\partial \mathbf{p_s}^2} $   & $ \dfrac{\partial^2 f(\mathbf{I_0}(\mathbf{p}), \mathbf{I_t})}{\partial \mathbf{p}^2} $
    \\[2ex]
    \color{blue}11 & $ \dfrac{\partial \mathbf{I_t}}{\partial \mathbf{p_s}} $,$ \dfrac{\partial^2 \mathbf{I_t}}{\partial \mathbf{p_s}^2} $      & $ \dfrac{\partial^2 f(\mathbf{I_0}, \mathbf{I_t}(\mathbf{p}))}{\partial \mathbf{p}^2} $
    \\[2ex]
    \color{blue}12 & $ \dfrac{\partial \mathbf{I_t}}{\partial \mathbf{p_s}} $,$ \dfrac{\partial^2 \mathbf{I_t}}{\partial \mathbf{p_s}^2} $      & $ \dfrac{\partial^2 f(\mathbf{I_t}, \mathbf{I_t}(\mathbf{p}))}{\partial \mathbf{p}^2} $  (Eq.  \ref{eq_hess_iclk}, \ref{eq_hess_fclk})
    \\[2ex]
    \color{blue}13 & $ \dfrac{\partial \mathbf{I_0}}{\partial \mathbf{p_s}} $,$ \dfrac{\partial^2 \mathbf{I_0}}{\partial \mathbf{p_s}^2} $, $ \dfrac{\partial \mathbf{I_t}}{\partial \mathbf{p_s}} $,$ \dfrac{\partial^2 \mathbf{I_t}}{\partial \mathbf{p_s}^2} $      & $ \dfrac{\partial^2 f(\mathbf{I_0}(\mathbf{p}), \mathbf{I_t})}{\partial \mathbf{p}^2}  + \dfrac{\partial^2 f(\mathbf{I_0}, \mathbf{I_t}(\mathbf{p}))}{\partial \mathbf{p}^2}$
    \\[2ex]
     \hline
    \end{tabular}
    \begin{tablenotes}
    \item[1] Functions {\color{blue}{10}}-{\color{blue}{13}} have overloaded variants that omit the second term in Eq. \ref{eq_hess_basic}, as  in Eq. \ref{eq_hess_basic_gn}, and so do not require $ \dfrac{\partial^2 \mathbf{I}}{\partial \mathbf{p_s}^2} $ as input
    \end{tablenotes}
\end{threeparttable}
\end{table}

\begin{table}[t]
\scriptsize
\centering
\begin{threeparttable}
    \caption{Specifications for important methods in \texttt{ILM}.}
    \label{tab_ilm_func_specification}   
    \begin{tabular}{|p{2.5cm}|l|p{3.0cm}|}	
     \hline
    \rowcolor{gray!50}
    \textbf{Function} & \textbf{Inputs} & \textbf{Output} \\
     \texttt{update} & $ \mathbf{p_{a}}, \Delta\mathbf{p_a} $      & $ \mathbf{p_{a}'}\ |\ \mathbf{g}(\mathbf{g}(\mathbf{I_t}, \mathbf{p_a}), \Delta\mathbf{p_a}) = \mathbf{g}(\mathbf{I_t}, \mathbf{p_a'})  $
     \\ 
    \texttt{apply} &  $ \mathbf{I_t}, \mathbf{p_a} $     &  $ \mathbf{g} ( \mathbf{I}, \mathbf{p_a}) $
    \\ 
    \texttt{invert} & $ \mathbf{p_a} $     &  $\mathbf{p_a'}\ |\ \mathbf{g}(\mathbf{g}(\mathbf{I_t}, \mathbf{p_a}), \mathbf{p_a'}) = \mathbf{I_t} $
    \\ 
    \texttt{cmptParamJacobian} & $ \dfrac{\partial f}{\partial \mathbf{g}}, \mathbf{I_t}, \mathbf{p_a} $      & $ \dfrac{\partial f}{\partial \mathbf{p_a}}$ 
    \\[2ex]
    \texttt{cmptPixJacobian} & $ \dfrac{\partial f}{\partial \mathbf{g}}, \mathbf{I_t}, \mathbf{p_a}$      &  $  \dfrac{\partial f}{\partial \mathbf{I_t}}$ 
    \\[2ex]
    \texttt{cmptParamHessian}\tnote{*} & $ \dfrac{\partial^2 f}{\partial \mathbf{g}^2}, \dfrac{\partial f}{\partial \mathbf{g}}, \mathbf{I_t}, \mathbf{p_a} $    & $ \dfrac{\partial^2 f}{\partial \mathbf{p_a}^2}$
     \\[2ex]
   \texttt{cmptPixHessian}\tnote{*} & $ \dfrac{\partial^2 f}{\partial \mathbf{g}^2}, \dfrac{\partial f}{\partial \mathbf{g}}, \mathbf{I_t}, \mathbf{p_a} $     & $ \dfrac{\partial^2 f}{\partial \mathbf{I_t}^2}$
    \\[2ex]
    \texttt{cmptCrossHessian}\tnote{*} & $ \dfrac{\partial^2 f}{\partial \mathbf{g}^2}, \dfrac{\partial f}{\partial \mathbf{g}}, \mathbf{I_t}, \mathbf{p_a} $      & $ \dfrac{\partial^2 f}{\partial \mathbf{I_t}\partial \mathbf{p_a}}$
    \\[2ex]
     \hline
    \end{tabular}
    \begin{tablenotes}
    \item[*] All Hessian functions have overloaded variants that omit the second terms in respective expressions and so do not require $ \dfrac{\partial f}{\partial \mathbf{g}} $ as input
    \end{tablenotes}
\end{threeparttable}
\end{table}
\subsubsection{Image Operations}
This part, abstracted into a separate class called \texttt{ImageBase},  handles all pixel level operations on
$ I_t $ like extracting the patch $ \mathbf{I}(\mathbf{x}) $ and computing its numerical gradient
$ \nabla \mathbf{I} $
and Hessian
$ \nabla^2 \mathbf{I} $.
It uses sub pixel interpolation so that these quantities can be treated as continuous functions of $ \mathbf{x} $.

Though \texttt{AM} bears a composition or "has a" relationship with \texttt{ImageBase},
the latter is actually implemented as a base class of the former to maintain simplicity of the interface and allow a specializing class to efficiently override functions in both classes. 
Moreover, having a separate class for pixel related operations means that AMs like SCV and ZNCC that differ from SSD only in using a modified version of $ \mathbf{I_0}$ or $\mathbf{I_t} $ (thus deriving from \texttt{SSDBase} in Fig. \ref{fig_class_diagram}), can implement the corresponding mapping entirely within the functions defined in \texttt{ImageBase} and be combined easily with other AMs besides SSD.

\subsubsection{Similarity Functions}
This is the core of \texttt{AM} and handles the computation of the similarity measure $ f(\mathbf{I^*}, \mathbf{I^c}, \mathbf{p_a}) $ and its derivatives $ \partial f/\partial \mathbf{I} $ and $ \partial^2 f/\partial \mathbf{I}^2 $ w.r.t. both $ \mathbf{I^*} $ and $ \mathbf{I^c}$. It also provides interfacing functions to use inputs from \texttt{SSM} to compute the derivatives of $ f $ w.r.t. SSM parameters using the chain rule. As a notational convention, all interfacing functions, including those in \texttt{SSM}, are prefixed with \texttt{cmpt}.

The functionality specific to $ \mathbf{p_a} $ is abstracted into a separate class called
\texttt{IlluminationModel} so it can be combined with any AM to add photometric parameters \citep{Silveira2007_esm_lighting, Bartoli2008} to it. This class provides functions to compute $ g(\mathbf{I},\mathbf{p_a}) $ and its derivatives including $ \partial g/\partial \mathbf{p_a} $, $ \partial^2 g/\partial \mathbf{p_a}^2 $, $ \partial g/\partial \mathbf{I} $, $ \partial^2 g/\partial \mathbf{I}^2 $ and $ \partial^2 g/\partial \mathbf{I}\partial \mathbf{p_a} $. These are called from within \texttt{AM} to compute the respective derivatives w.r.t. $ f $ so that the concept of ILM is transparent to the SM. 
It should be noted that \texttt{AM} is designed to support $ f $ with arbitrary $ \mathbf{p_a}$ of which  ILM is a special case. It also supports learning to update the object's appearance, as present, for instance, in \texttt{PCA} \citep{Ross08ivt}.

Since several of the functions in this part of \texttt{AM} involve common computations, there exist \textit{transitive dependency} relationships between them (Fig. \ref{fig_am_dependency}) to avoid repeating these computations when multiple quantities are needed by the SM. What this means is that a function lower down in the dependency hierarchy may delegate part of its computations to any function higher up in the hierarchy so that the latter must be called \textit{before} calling the former if correct results are to be expected.

\subsubsection{Distance Feature}
\label{sec_dist_feature}
This part, implemented within \texttt{AMDist}, is designed specifically to enable integration with the FLANN library \citep{Muja2009_flann} that is used by the NN based SM. It provides two main functions: 
\begin{itemize}
\item A feature transform $ \mathbf{D}(\mathbf{I^*})  : \mathbb R^N\mapsto \mathbb R^K $   that maps the pixel values extracted from a patch $ \mathbf{I^*} $ into a feature vector that contains the results of all computations in $ f (\mathbf{I^*}, \mathbf{I^c}) $ that depend only on $ \mathbf{I^*} $, and likewise for $ \mathbf{I^c} $. This transform is applied to all sampled patches during initialization and only the resultant feature vectors are stored in the index. At runtime, it is applied to $ \mathbf{I^c} $ and the feature vector is passed to the distance functor.

\item A highly optimized distance functor $f_D( \mathbf{D}(\mathbf{I^*}), \mathbf{D}(\mathbf{I^c})) : \mathbb{R}^K \times \mathbb{R}^K \mapsto \mathbb{R}$ that computes a measure of the distance or dissimilarity between $ \mathbf{I^*} $ and $ \mathbf{I^c} $ (typically the negative of $ f(\mathbf{I^*}, \mathbf{I^c}) $) given the distance features $ \mathbf{D}(\mathbf{I^*}) $ and  $\mathbf{D}(\mathbf{I^c}) $ as inputs. 
\end{itemize}
The main idea behind the design of these two components is to place as much computational load as possible on $ \mathbf{D} $ so that the runtime speed of $f_D $ is maximized, with the premise that the former is called mostly during initialization when the sample dataset is to be built, while the latter is called online to find the best matches for a candidate patch in the dataset.
An optimal design may involve a trade off between the size $ K $ of the feature vector and the amount of computation performed in $f_D$.
For non symmetrical AMs, i.e. where $ f(\mathbf{I^*}, \mathbf{I^c}) \neq f(\mathbf{I^c}, \mathbf{I^*})$ (e.g. CCRE and SCV), the feature vector may also include an indicator flag so that $ f_D $ can determine which of its arguments corresponds to the $ \mathbf{D}(\mathbf{I^*}) $ and which to $ \mathbf{D}(\mathbf{I^c}) $. This is needed because FLANN does not specify the order in which the arguments will be passed to $ f_D $ and examination of its code showed that this order varies for each index type as well as for different calls to $ f_D $ within the same index.

\subsection{\em{\texttt{StateSpaceModel}}}
\label{sec_ssm}
\setlength{\tabcolsep}{4pt}
\begin{table}[t]
	\centering
	\caption{Specifications for important methods in \texttt{SSM}.}
	\label{tab_ssm_func_specification}   
	\begin{tabular}{|p{3.0cm}|l|p{3.3cm}|}	
		\hline
		\rowcolor{gray!50}
		Function & Inputs & Output/Result \\
		\scriptsize{\texttt{compositionalUpdate}} & \scriptsize{$ \Delta\mathbf{p_s} $}      & \scriptsize{$ \mathbf{p_{st}}=\mathbf{p_s}'\ |\ \mathbf{w}(\mathbf{x}, \mathbf{p_s}')= \mathbf{w}(\mathbf{w}(\mathbf{x}, \Delta \mathbf{p_s}), \mathbf{p_{st}})  $}
		\\ 
		\scriptsize{\texttt{additiveUpdate}} &  \scriptsize{$ \Delta\mathbf{p_s} $}      &  \scriptsize{$ \mathbf{p_{st}} = \mathbf{p_{st}} +  \Delta\mathbf{p_s} $}
		\\ 
		\scriptsize{\texttt{invertState}} & \scriptsize{$ \mathbf{p_s} $ }     &  \scriptsize{$\mathbf{p_s'}\ |\ \mathbf{w}(\mathbf{w}(\mathbf{x}, \mathbf{p_s}), \mathbf{p_s'}) = \mathbf{x} $ }
		\\ 
		\scriptsize{\texttt{cmptPixJacobian}} & \scriptsize{$ \nabla\mathbf{I_t} $}      & \scriptsize{$ \left.\dfrac{\partial\mathbf{I_t}}{\partial\mathbf{p_s}} \right\rvert_{\mathbf{p_s}=\mathbf{p_{st}}}$} (Eq. \ref{eq_jac_falk})
		\\[2ex]
		\scriptsize{\texttt{cmptWarpedPixJacobian}} & \scriptsize{$ \nabla\mathbf{I_t}$}      &  \scriptsize{$  \left.
			\dfrac{\partial\mathbf{I_t}(\mathbf{w})}{\partial \mathbf{p_s}}
			\right\rvert_{\mathbf{p_s}=\mathbf{p_{s0}}}
			$} (Eq. \ref{eq_jac_fclk}, \ref{eq_jac_warped_grad})
		\\[2ex]
		\scriptsize{\texttt{cmptApproxPixJacobian}} & \scriptsize{$ \nabla\mathbf{I_0} $}      & \scriptsize{
			$ \dfrac{\partial \mathbf{I_t}}{\partial \mathbf{p_{st}}}$ (approx)
		} (Eq. \ref{eq_jac_ialk}, \ref{eq_ialk_assumption})
		\\[2ex]
		\scriptsize{\texttt{cmptPixHessian}} &   \scriptsize{$ \nabla\mathbf{I_t},\nabla^2\mathbf{I_t}  $}      & \scriptsize{$ 
			\left.
			\dfrac{\partial^2 \mathbf{\mathbf{I_t}}}{\partial \mathbf{p_s}^2}
			\right\rvert_{\mathbf{p_s}=\mathbf{p_{st}}}
			$  (Eq. \ref{eq_pix_hess_falk})}
		\\[2ex]
		\scriptsize{\texttt{cmptWarpedPixHessian}} &  \scriptsize{$\nabla\mathbf{I_t},\nabla^2\mathbf{I_t} $}      & \scriptsize{$
			\left.
			\dfrac{\partial^2 \mathbf{\mathbf{I_t}(\mathbf{w})}}{\partial \mathbf{p_s}^2}
			\right\rvert_{\mathbf{p_s}=\mathbf{p_{s0}}}
			$(Eq. \ref{eq_pix_hess_fclk}, \ref{eq_warped_pix_hess})}
		\\[2ex]
		\scriptsize{\texttt{cmptApproxPixHessian}} & \scriptsize{$\nabla\mathbf{I_0},\nabla^2\mathbf{I_0} $}      & \scriptsize{$  \dfrac{\partial^2 \mathbf{\mathbf{I_t}}}{\partial \mathbf{p_{st}}^2} $ (approx) (Eq. \ref{eq_ialk_pix_hess}) }
		\\[2ex]
		\hline
	\end{tabular}
\end{table}
This class has a simpler internal state than \texttt{AM} and can be described by only three main variables at any time $ t $ - sampled grid points $ \mathbf{x_t} $, corresponding corners $ \mathbf{x^c_t} $ and state parameters $ \mathbf{p_{st}} $.
It may be noted (Fig. \ref{fig_class_diagram}) that, though \texttt{SSM} is designed to support any arbitrary $ \mathbf{w} $,
most SSMs currently implemented are subsets of the planar projective transform and so derive from \texttt{ProjectiveBase} that abstracts out the functionality common to these.

Functions in \texttt{SSM} can be divided into two
categories:
\subsubsection{Warping Functions}
\label{sec_ssm_warping_function}
This is the core of \texttt{SSM} and provides a function $ \mathbf{w} $ to transform a regularly spaced grid of points $ \mathbf{x_0} $ representing the target patch into a warped patch $ \mathbf{x_t} = \mathbf{w}(\mathbf{x_0}, \mathbf{p_{st}}) $ that captures the tracked object's motion in image space. 
It also allows for the compositional inverse of $ \mathbf{w} $ to be computed (\texttt{\small{invertState}}) to support inverse SMs.
Further, there are functions to compute the derivatives of $ \mathbf{w} $ w.r.t. both $ \mathbf{x} $ and $ \mathbf{p_s}$
but, unlike \texttt{AM}, \texttt{SSM}
does not
store these as state variables,
rather their computation is implicit in the interfacing functions that compute
$ \partial \mathbf{I}/\partial \mathbf{p_s} $ and $ \partial^2 \mathbf{I}/\partial \mathbf{p_s}^2 $
using the chain rule.
This design decision was made for reasons of efficiency since $ \partial\mathbf{w}/\partial \mathbf{p_s} $ and $\partial\mathbf{w}/\partial \mathbf{x}$ are large and often very sparse tensors.
Computing these separately, thus, not only wastes a lot of memory but is also very inefficient.

Finally, there are four ways to update the internal state: incrementally using additive (\texttt{\small{additiveUpdate}}) or compositional (\texttt{\small{compositionalUpdate}}) formulations, or outright by providing either the state vector (\texttt{\small{setState}}) or the corresponding corners (\texttt{\small{setCorners}}) that define the current location of the patch.
There are no complex dependencies in \texttt{SSM} - the correct performance of interfacing functions and accessors depends only on one of the update functions being called every iteration. Table \ref{tab_ssm_func_specification} lists the functionality of some important methods in this part.

\subsubsection{Stochastic Sampler}
\label{sec_ssm_stochastic_sampler}
This part is provided to support stochastic SMs
and offers following functionality to this end:
\begin{itemize}
\item generate small random incremental updates to $ \mathbf{p_s} $ (\texttt{\small{generatePerturbation}}) by drawing these from a zero mean normal distribution with either user provided or heuristically estimated (\texttt{\small{estimateStateSigma}}) variance. 
\item generate stochastic state samples using the given state transition model - currently random walk (\texttt{\small{additiveRandomWalk}}) and first order auto regression (\texttt{\small{additiveAutoRegression1}}) are supported. There are also \texttt{\small{compositional}} variants.
\item estimate the mean of a set of samples of $ \mathbf{p_s} $ (\texttt{\small{estimateMeanOfSamples}})
\item estimate the best fit $ \mathbf{p_s} $ from a set of original and warped point pairs (\texttt{\small{estimateWarpFromPts}}) using a robust method implemented within \texttt{\small{SSMEstimator}} - currently RANSAC \citep{Zhang2015_rklt} and LMS \citep{Rousseeuw84_lmeds} are supported.
\end{itemize}

%% file: examples.tex
\section{Search Methods}
\label{examples}
\setlength{\floatsep}{0pt}

This section presents pseudo codes for several SMs currently implemented in MTF
to exemplify the usage of functions described in the previous sections. 
Algorithms 1-5 illustrate how MTF implements the many variants of gradient based SMs developed over the past 35 years \citep{Baker04lucasKanade_paper} while
Algorithms 6-8 illustrate stochastic SMs (Sec. \ref{stochastic_search}).
Following are some points and conventions to be noted:
\begin{itemize}
\item different algorithms make extensive references to portions of each other not only to
avoid redundancy but also to emphasize the parts they have in common.
\item \textit{am} and \textit{ssm} respectively refer to instances of \texttt{\small{AM}} and \texttt{\small{SSM}} (or rather of specializations thereof)
\item \textit{am} has direct access to the latest image in the sequence so it is not passed explicitly in function calls - this is one of the design features of \texttt{\small{AM}} to avoid the overhead of passing the image repeatedly.
\item several special cases like the optional use of the first order Hessian (Eq. \ref{eq_hess_basic_gn}), parameterization and online learning of AM and iterative form of the \texttt{\small{update}} function are demonstrated only in Alg. \ref{alg_iclk} but should be obvious by analogy for other SMs too.
\item \textit{v}.head(\textit{h}) and \textit{v}.tail(\textit{t}) in Alg. \ref{alg_iclk} respectively refer to the first \textit{h} and last \textit{t} elements in the $ \textit{h} + \textit{t} $ length vector $ \textit{v} $.
\item \textit{flann} in Alg. \ref{alg_nn} is an instance of FLANN library \citep{Muja2009_flann} that can build an index from a set of samples and search it for a new candidate.
\item variables used to store the results of computations are not described explicitly but their meanings should be clear from their names and context.
For instance, \textit{sample\_dataset} and \textit{ssm\_perturbations} used in Alg. \ref{alg_nn} respectively refer to $ n{\times} K $ and $ n{\times} S $ matrices, each of whose rows contains the distance feature $ \mathbf{D} $ (Sec. \ref{sec_dist_feature}) and the SSM state $ \mathbf{p_s} $ corresponding to one sample so that $ n = $ number of samples.
\item only one state transition model is shown in Alg. \ref{alg_pf} though several others are available too (Sec. \ref{sec_ssm_stochastic_sampler}).
\item  Alg. \ref{alg_lms_ransac} shows only the \texttt{\small{GridTracker}} component of LMS and RANSAC; the actual robust estimation using one of these is performed in \texttt{\small{estimateWarpFromPts}}
function of \texttt{\small{SSM}}
(Sec. \ref{sec_ssm_stochastic_sampler}).
\item  sampling resolution of \textit{ssm} in  Alg. \ref{alg_lms_ransac} is set to be same as the grid resolution and the function \texttt{\small{getRegion}} (\textit{c}, \textit{s}) in line 6 returns the corners of a rectangular region of size \textit{s} with centroid \textit{c}.
\end{itemize}

\begin{algorithm}[t]
\scriptsize
\caption{ICLK}
\label{alg_iclk}
\begin{algorithmic}[1]
\Function {initialize}{$ \textit{corners} $}
\State
$ \textit{ssm} $.initialize($ \textit{corners} $)
\State
$ \textit{am} $.initializePixVals($ \textit{ssm} $.getPts())
\State
$ \textit{am} $.initializePixGrad($ \textit{ssm} $.getPts())
\State
$ \textit{am} $.initializeSimilarity()
\State
$ \textit{am} $.initializeGrad()
\State
$ \textit{am} $.initializeHess()
\State
$ \textit{dI0\_dps}\gets$ $ \textit{ssm} $.cmptWarpedPixJacobian($ \textit{am} $.getInitPixGrad())
\If{ \textit{use\_first\_order\_hessian}}
\State
$\textit{d2f\_dp2} \gets\textit{am}$.cmptSelfHessian($ \textit{dI0\_dps} $)
\Else
\State
$ \textit{am} $.initializePixHess($ \textit{ssm} $.getPts())
\State
$\textit{d2I0\_dps2}\gets \textit{ssm} $.cmptInitPixHessian(
\Statex
\-\hspace{2cm}$\textit{am} $.getInitPixHess(), $\textit{am} $.getInitPixGrad())
\State
$ \textit{d2f\_dp2} $ $\gets\textit{am}$.cmptSelfHessian($ \textit{dI0\_dps} $, $ \textit{d2I0\_dps2} $)
\EndIf
\EndFunction
\Function {update}{}
\For{$i\gets 1, \textit{max\_iters}$}
\State
\textit{am}.updatePixVals(\textit{ssm}.getPts())
\State
\textit{am}.updateSimilarity()
\State
\textit{am}.updateInitGrad()
\State
\textit{df\_dp} $ \gets $ \textit{am}.cmptInitJacobian($ \textit{dI0\_dps} $)
\State
\textit{delta\_p} $ \gets - \textit{d2f\_dp2}$.inverse()$* \textit{df\_dp}$
\State
\textit{delta\_ps} $ \gets $ \textit{delta\_p}.head(\textit{ssm}.getStateSize())
\State
\textit{delta\_pa} $ \gets $ \textit{delta\_p}.tail(\textit{am}.getStateSize())
\State
\textit{inv\_delta\_ps} $ \gets $\textit{ssm}.invertState(\textit{delta\_ps})
\State
\textit{inv\_delta\_pa} $ \gets $\textit{am}.invertState(\textit{delta\_pa})
\State
\textit{prev\_corners}$ \gets $\textit{ssm}.getCorners()
\State
\textit{ssm}.compositionalUpdate(\textit{inv\_delta\_ps})
\State
\textit{am}.update(\textit{inv\_delta\_pa})
\If{ $ \| \textit{prev\_corners} - \textit{ssm}.\text{getCorners}()\|^2 < \epsilon $}
\State
\textbf{break}
\EndIf
\EndFor
\State
\textit{am}.updateModel(\textit{ssm}.getPts())
\State
\textbf{return} \textit{ssm}.getCorners()
\EndFunction
\end{algorithmic}
\end{algorithm}

\begin{algorithm}[!htbp]
\scriptsize
\caption{FCLK}
\label{alg_fclk}
\begin{algorithmic}[1]
\Function {initialize}{$ \textit{corners} $}
\State
lines 2-7 of Alg. \ref{alg_iclk}
\State
$ \textit{am} $.initializePixHess($ \textit{ssm} $.getPts())
\EndFunction
\Function {update}{}
\State
lines 19-20 of Alg. \ref{alg_iclk}
\State
\textit{am}.updateCurrGrad()
\State
\textit{am}.updatePixGrad(\textit{ssm}.getPts())
\State
\textit{am}.updatePixHess($ \textit{ssm} $.getPts())
\State
\textit{dIt\_dps} $\gets$ $ \textit{ssm} $.cmptWarpedPixJacobian($ \textit{am} $.getCurrPixGrad())
\State
\textit{d2It\_dps2} $\gets$ $ \textit{ssm} $.cmptWarpedPixHessian(
\Statex
\-\hspace{2cm}$\textit{am} $.getCurrPixHess(), $\textit{am} $.getCurrPixGrad())
\State
\textit{df\_dp} $ \gets $\textit{am}.cmptCurrJacobian(\textit{dIt\_dps})
\State
\textit{d2f\_dp2} $\gets$\textit{am}.cmptSelfHessian(\textit{dIt\_dps}, \textit{d2It\_dps2})
\State
\textit{delta\_p} $ \gets - \textit{d2f\_dp2}$.inverse()$* \textit{df\_dp}$
\State
\textit{ssm}.compositionalUpdate(\textit{delta\_p})
\State
\textbf{return} \textit{ssm}.getCorners()
\EndFunction
\end{algorithmic}
\end{algorithm}

\begin{algorithm}[t]
\scriptsize
\caption{ESM}
\label{alg_esm}
\begin{algorithmic}[1]
\Function {initialize}{$ \textit{corners} $}
\State
lines 2-3 of Alg. \ref{alg_fclk}
\State
$ \textit{dI0\_dps} \gets$ $ \textit{ssm} $.cmptWarpedPixJacobian($ \textit{am} $.getInitPixGrad())
\State
\textit{d2f\_dp2\_0} $\gets$ \textit{am}.cmptSelfHessian(\textit{dI0\_dps}, \textit{d2I0\_dps2})
\EndFunction
\Function {update}{}
\State
lines 6-11 of Alg. \ref{alg_fclk}
\State
\textit{am}.updateInitGrad()
\State
\textit{df\_dp}$ \gets $\textit{am}.cmptDifferenceOfJacobians(\textit{dI0\_dps}, \textit{dIt\_dps})
\State
\textit{d2f\_dp2\_t} $\gets$\textit{am}.cmptSelfHessian(\textit{dIt\_dps}, \textit{d2It\_dps2})
\State
\textit{d2f\_dp2} $\gets$$  \textit{d2f\_dp2\_0} + \textit{d2f\_dp2\_t} $
\State
lines 14-16 of Alg. \ref{alg_fclk}
\EndFunction
\end{algorithmic}
\end{algorithm}

\begin{algorithm}[!htbp]
\scriptsize
\caption{IALK}
\label{alg_ialk}
\begin{algorithmic}[1]
\Function {initialize}{$ \textit{corners} $}
\State
same as Alg. \ref{alg_fclk}
\EndFunction
\Function {update}{}
\State
lines 6-7 of Alg. \ref{alg_fclk}
\State
$ \textit{dIt\_dps} \gets$ $ \textit{ssm} $.cmptApproxPixJacobian($ \textit{am} $.getInitPixGrad())
\State
$ \textit{d2It\_dps2} \gets$ $ \textit{ssm} $.cmptApproxPixHessian($\textit{am} $.getInitPixHess(), 
\Statex
\-\hspace{2cm}$\textit{am} $.getInitPixGrad())
\State
lines 12-14 of Alg. \ref{alg_fclk}
\State
\textit{ssm}.additiveUpdate(\textit{delta\_p})
\State
\textbf{return} \textit{ssm}.getCorners()
\EndFunction
\end{algorithmic}
\end{algorithm}

\begin{algorithm}[!htbp]
\scriptsize
\caption{FALK}
\label{alg_falk}
\begin{algorithmic}[1]
\Function {initialize}{$ \textit{corners} $}
\State
same as Alg. \ref{alg_fclk}
\EndFunction
\Function {update}{}
\State
lines 6-9 of Alg. \ref{alg_fclk}
\State
$ \textit{dIt\_dps} \gets$ $ \textit{ssm} $.cmptPixJacobian($ \textit{am} $.getCurrPixGrad())
\State
$ \textit{d2It\_dps2} \gets$ $ \textit{ssm} $.cmptPixHessian($\textit{am} $.getCurrPixHess(), 
\Statex
\-\hspace{2cm}$\textit{am} $.getCurrPixGrad())
\State
lines 8-10 of Alg. \ref{alg_ialk}
\EndFunction
\end{algorithmic}
\end{algorithm}

\begin{algorithm}[t]
\scriptsize
\caption{NN}
\label{alg_nn}
\begin{algorithmic}[1]
\Function {initialize}{$ \textit{corners} $}
\State
lines 2-3 of Alg. \ref{alg_iclk}
\State
\textit{state\_sigma}$ \gets $ \textit{ssm}.estimateStateSigma()
\State
\textit{ssm}.initializeSampler(\textit{state\_sigma})
\State
\textit{am}.initializeDistFeat()
\For{\textit{sample\_id} $ \gets 1, $ \textit{no\_of\_samples}}
\State
\textit{ssm\_updates}.row(\textit{sample\_id}) $ \gets $ \textit{ssm}.generatePerturbation()
\State
\textit{inv\_update} $\gets $\textit{ssm}.invertState(\textit{ssm\_updates}.row(\textit{sample\_id}))
\State
\textit{ssm}.compositionalUpdate(\textit{inv\_update})
\State
\textit{am}.updatePixVals(\textit{ss}m.getPts())
\State
\textit{am}.updateDistFeat()
\State
\textit{sample\_dataset}.row(\textit{sample\_id}) $\gets$ \textit{am}.getDistFeat()
\State
\textit{ssm}.compositionalUpdate(\textit{ssm\_updates}.row(\textit{sample\_id}))
\EndFor
\State
\textit{flann}.buildIndex(\textit{sample\_dataset})
\EndFunction

\Function {update}{}
\State
\textit{am}.updatePixVals(\textit{ssm}.getPts())
\State
\textit{am}.updateDistFeat()
\State
\textit{nn\_sample\_id} $ \gets $ \textit{flann}.searchIndex(\textit{am}.getDistFeat())
\State
\textit{ssm}.compositionalUpdate(\textit{ssm\_updates}.row(\textit{nn\_sample\_id}))
\State
\textbf{return} \textit{ssm}.getCorners()
\EndFunction
\end{algorithmic}
\end{algorithm}

\begin{algorithm}[!htbp]
\scriptsize
\caption{PF}
\label{alg_pf}
\begin{algorithmic}[1]
\Function {initialize}{$ \textit{corners} $}
\State
lines 2-4 of Alg. \ref{alg_nn}
\State
\textit{am}.initializeSimilarity()
\For{\textit{particle\_id} $ \gets 1, $ \textit{no\_of\_particles}}
\State
\textit{particles}[\textit{particle\_id}].\textit{state} $ \gets $ \textit{ssm}.getState()
\State
\textit{particles}[\textit{particle\_id}].\textit{weight} $ \gets $ $ 1/\textit{no\_of\_particles} $
\EndFor
\EndFunction

\Function {update}{}
\For{\textit{particle\_id} $ \gets 1, $ \textit{no\_of\_particles}}
\State
\textit{particles}[\textit{particle\_id}].\textit{state} $\gets$ \textit{ssm}.compositionalRandomWalk(
\State
\-\hspace{2cm}\textit{particles}[\textit{particle\_id}].\textit{state})
\State
\textit{ssm}.setState(\textit{particles}[\textit{particle\_id}].\textit{state})
\State
\textit{am}.updatePixVals(\textit{ssm}.getPts())
\State
\textit{am}.updateSimilarity()
\State
\textit{particles}[\textit{particle\_id}].\textit{weight} $\gets $\textit{am}.getLikelihood()
\EndFor
\State
normalize weights and resample the particles
\State
\textit{mean\_state} $ \gets $ \textit{ssm}.estimateMeanOfSamples(\textit{particles});
\State
\textit{ssm}.setState(\textit{mean\_state})
\State
\textbf{return} \textit{ssm}.getCorners()
\EndFunction
\end{algorithmic}
\end{algorithm}

\begin{algorithm}[!htbp]
\scriptsize
\caption{LMS/RANSAC}
\label{alg_lms_ransac}
\begin{algorithmic}[1]
\Function {initialize}{$ \textit{corners} $}
\State
\textit{sub\_trackers} $ \gets $ vector of 2 DOF sub patch trackers
\State
\textit{ssm}.initialize($ \textit{corners} $)
\State
\textit{curr\_pts} $ \gets $ \textit{ssm}.getPts()
\For{\textit{pt\_id} $ \gets 1, $ \textit{no\_of\_pts}}
\State
\textit{sub\_patch\_corners} $ \gets $ getRegion(\textit{curr\_pts}[\textit{pt\_id}], \textit{sub\_patch\_size})
\State
\textit{sub\_trackers}[\textit{pt\_id}].initialize(\textit{sub\_patch\_corners})
\EndFor
\EndFunction

\Function {update}{}
\State
\textit{prev\_pts} $ \gets $ \textit{curr\_pts}
\For{\textit{pt\_id} $ \gets 1, $ \textit{no\_of\_pts}}
\State
\textit{sub\_trackers}[\textit{pt\_id}].update()
\State
\textit{curr\_pts}[\textit{pt\_id}] $ \gets $ getCentroid(\textit{sub\_trackers}[\textit{pt\_id}].getRegion())
\EndFor
\State
\textit{opt\_warp} $ \gets $ \textit{ssm}.estimateWarpFromPts(\textit{prev\_pts}, \textit{curr\_pts})
\State
\textit{warped\_corners} $ \gets $ \textit{ssm}.applyWarpToCorners(\textit{ssm}.getCorners(), \textit{opt\_warp})
\State
\textit{ssm}.setCorners(\textit{warped\_corners})
\State
lines 4-8 of Alg. \ref{alg_lms_ransac}
\State
\textbf{return} \textit{ssm}.getCorners()
\EndFunction
\end{algorithmic}
\end{algorithm}

%% file: use_cases.tex
\section{Use Cases}
\label{use_cases}

\begin{algorithm}[!htbp]
\caption{Object Tracking - Simple}
\label{use_case_tracking}
\scriptsize
\begin{algorithmic}[1]
\State using namespace mtf;
\State \texttt{ICLK}$ < $\texttt{SSD}, \texttt{Homography}$ > $ \textit{tracker};
\State \texttt{GaussianSmoothing} \textit{pre\_proc}(\textit{input}.getFrame(), \textit{tracker}.inputType());
\State \textit{tracker}.initialize(\textit{pre\_proc}.getFrame(), \textit{init\_location});
\While{\textit{input}.update()}
\State \textit{pre\_proc}.update(\textit{input}.getFrame());
\State \textit{tracker}.update(\textit{pre\_proc}.getFrame());
\State \textit{new\_location} $ \gets $ \textit{tracker}.getRegion();
\EndWhile
\end{algorithmic}
\end{algorithm}

\begin{algorithm}[!htbp]
\caption{Object Tracking - Composite}
\label{use_case_tracking_composite}
\scriptsize
\begin{algorithmic}[1]
\State \texttt{PF}$ < $\texttt{ZNCC}, \texttt{Affine}$ > $ \textit{tracker1};
\State \texttt{FCLK}$ < $\texttt{SSIM}, \texttt{SL3}$ > $ \textit{tracker2};
\State vector$ < $\texttt{TrackerBase*}$ > $ \textit{trackers} = \{\&\textit{tracker1}, \&\textit{tracker2}\};
\State \texttt{\texttt{CascadeTracker}} \textit{tracker}(\textit{trackers});
\State lines 3-9 of Alg. \ref{use_case_tracking}
\end{algorithmic}
\end{algorithm}

\begin{algorithm}[!htbp]
\caption{UAV Trajectory Estimation in Satellite Image}
\label{use_case_uav}
\scriptsize
\begin{algorithmic}[1]
\State \texttt{ESM}$ < $\texttt{MI}, \texttt{Similitude}$ > $ \textit{tracker};
\State \textit{uav\_img\_corners}$ \gets $getFrameCorners(\textit{input}.getFrame());
\State \textit{tracker}.initialize(\textit{satellite\_img}, \textit{init\_uav\_location});
\State \textit{curr\_uav\_location}$ \gets $\textit{tracker}.getRegion();
\While{\textit{input}.update()}
\State \textit{tracker}.initialize(\textit{input}.getFrame(), \textit{uav\_img\_corners});
\State \textit{tracker}.setRegion(\textit{curr\_uav\_location});
\State \textit{tracker}.update(\textit{satellite\_img});
\State \textit{curr\_uav\_location}$ \gets $\textit{tracker}.getRegion();
\EndWhile
\end{algorithmic}
\end{algorithm}

\begin{algorithm}[!htbp]
\caption{Online Image Mosaicing}
\label{use_case_mosaic}
\scriptsize
\begin{algorithmic}[1]
\State \texttt{FALK}$ < $\texttt{MCNCC}, \texttt{Isometry}$ > $ \textit{tracker};
\State \textit{mos\_img} $ \gets $ writePixelsToImage(\textit{input}.getFrame(), \textit{init\_mos\_location}, \textit{mos\_size});
\State \textit{mos\_location} $ \gets $  \textit{init\_mos\_location};
\While{\textit{input}.update()}
\State \textit{temp\_img} $ \gets $ writePixelsToImage(\textit{input}.getFrame(), \textit{mos\_location}, \textit{mos\_size});
\State \textit{tracker}.initialize(\textit{temp\_img}, \textit{mos\_location});
\State \textit{tracker}.update(\textit{mos\_img});
\State \textit{mos\_location} $ \gets $ \textit{tracker}.getRegion();
\State \textit{mos\_img} $ \gets $ writePixelsToImage(\textit{input}.getFrame(), \textit{mos\_location}, \textit{mos\_size});
\EndWhile
\end{algorithmic}
\end{algorithm}

This section presents the following use cases for MTF in C++ style pseudo code:
\begin{itemize}
\item Track an object in an image sequence using a simple (Alg. \ref{use_case_tracking}) and a composite (Alg. \ref{use_case_tracking_composite}) tracker.
\item Estimate the trajectory of a UAV within a large satellite image of an area from images it took while flying over that area (Alg. \ref{use_case_uav}).
\item Create an image mosaic in real time from a video sequence captured by a camera moving over different parts of the planar scene to be stitched (Alg. \ref{use_case_mosaic}).
\end{itemize}
Visual illustrations for all of these can be found in the demonstrations video on MTF website \citep{mtfweb}.
Following are details regarding variables and functions used in these algorithms that assist the reader in understanding them better:
\begin{itemize}
\item MTF comes with an input module with wrappers for image capturing functions in OpenCV, ViSP and XVision.
It is represented here by \textit{input} and is assumed to have been initialized with the appropriate source.
\item Raw images acquired by the input module can optionally be passed to the preprocessing module that provides wrappers for OpenCV image filtering and conversion functions.
An example
of this
is
\texttt{\small{GaussianSmoothing}} in Alg. \ref{use_case_tracking}.
\item Though only five combinations of SM, AM and SSM are shown here, these can be replaced by virtually any combination of methods in Fig. \ref{fig_class_diagram}.
\item MTF also has a set of general utility functions for image and warping related operations. The following have been used in Alg. \ref{use_case_uav} and \ref{use_case_mosaic}:
\begin{itemize}
\item \texttt{\small{getFrameCorners}}(\textit{image}) returns a $ 2\times 4 $ matrix containing the corners of \textit{image} and is thus used when the entire image is to be considered as the tracked region.
\item \texttt{\small{writePixelsToImage}}(\textit{patch}, \textit{corners}, \textit{size}) writes the pixel values in \textit{patch} to an image with dimensions \textit{size} within the region bounded by \textit{corners}.
\end{itemize}
\item \textit{init\_mos\_location} in Alg. \ref{use_case_mosaic} is the user specified location of the first frame in the sequence within the mosaic image of size \textit{mos\_size}.
This is typically the center of the mosaic though can be elsewhere depending on the actual sequence.
\end{itemize}

%% file: performance.tex
\section{Performance}
\label{sec_performance}
\begin{figure}[t]
	\begin{center}
		\includegraphics[width=0.5\textwidth]{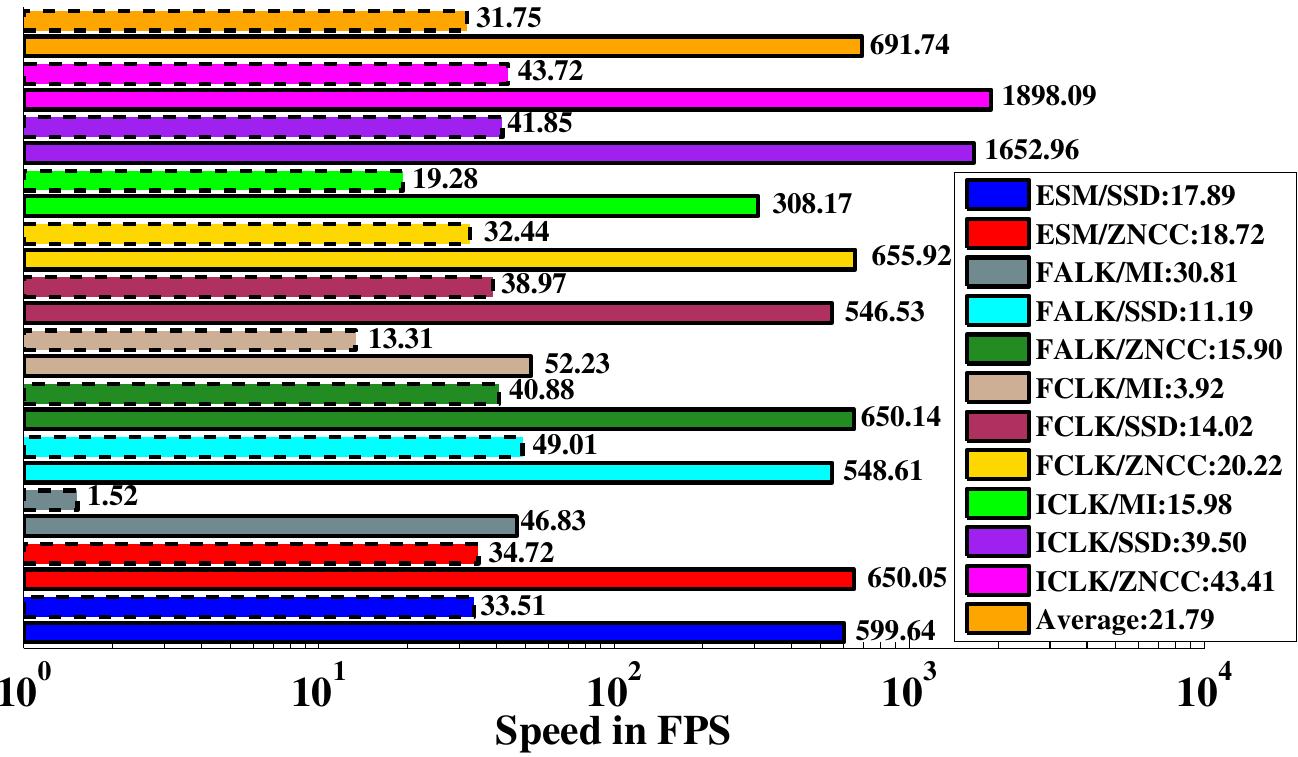}\\
		\caption{ViSP vs MTF average tracker speeds for all combinations of SMs and AMs supported by ViSP. MTF and ViSP results are shown in \textbf{solid and dotted lines} respectively.
			Note that \textbf{logarithmic scaling} has been used on the x axis for better visibility of ViSP bars though the actual figures are also shown.
			\textbf{Speedup} provided by MTF is shown in the legends.
			Results were generated on a 4 GHz Intel Core i7-4790K machine with 32 GB RAM.
		}
		\label{fig_vp_mtf_speed_hom}
	\end{center}
\end{figure}
\begin{figure*}[!htbp]
	\begin{center}
		\setlength{\belowcaptionskip}{-10pt}
		\includegraphics[width=\textwidth]{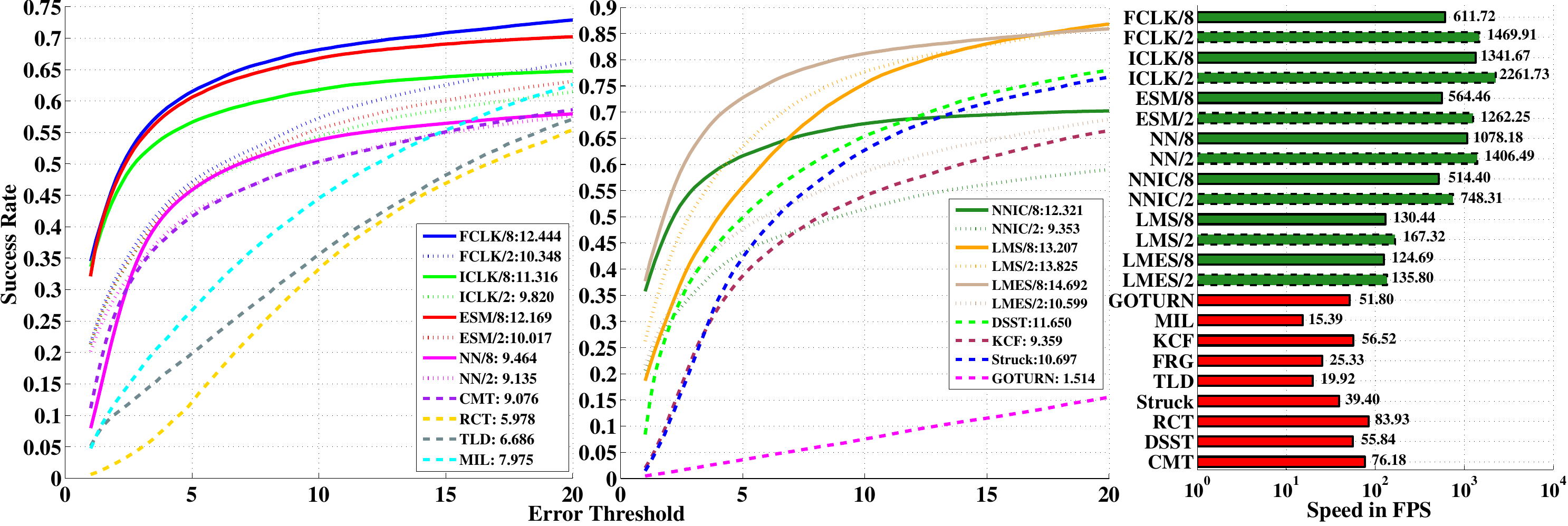}\\
		\caption{Comparing OLTs with 2 and 8 DOF RBTs in terms of both accuracy and speed.
			OLTs are shown in \textbf{dashed} lines while 2 and 8 DOF RBTs are in \textbf{dotted} and \textbf{solid} lines respectively.
			Accuracy is measured by the tracking success rates (\textbf{SR}) for a range of error thresholds ($ t_p $) over 4 large datasets with more than 100000 frames.
			OLTs and 2 DOF RBTs were evaluated against 2 DOF ground truth for fairness.
			Legends in the SR plots show the areas under the respective curves.
			Speed plot on the right has \textbf{logarithmic scaling} on the x axis for clarity though actual figures are also shown.
			Original C++ implementations were used for all OLTs.
			More details about the evaluation methodology and the meanings of several acronyms can be found in \citep{singh17_ssim, singh17_mtf_thesis}.
		}
		\label{fig_olt_vs_rbt_2_8}
	\end{center}
\end{figure*}
An existing system that is functionally similar to MTF is the template tracker module of the ViSP library \citep{Marchand05_visp}.
It provides 4 SMs, 3 AMs and 6 SSMs though not all combinations work.
MTF offers several advantages over ViSP.
Firstly, SMs and AMs in ViSP are not implemented as independent modules; rather, each combination of methods has its own class. This makes it difficult to add a new method for either of these sub modules and combine it with existing methods for the others.  
Secondly, MTF has several more AMs, one more GD based SM as well as four stochastic SMs. It also allows multiple SMs to be combined effortlessly to create novel composite SMs. 
Thirdly, MTF is significantly faster than ViSP. As shown in fig. \ref{fig_vp_mtf_speed_hom}, MTF is usually more than an order of magnitude faster, with its speed being over $ 20 $ times higher than ViSP on average.
This is mainly because MTF uses the Eigen library \citep{eigenweb} for all mathematical computations and this is known to be one of the fastest \citep{eigen_benchmark}.
Lastly, ViSP trackers are significantly buggy and only completed about $  70\% $ of the tested sequences.
This section is concluded with a comparison of MTF trackers with 8 state of the art OLTs \citep{Kristan2016_vot16}
to validate the suitability of the former for robotics applications.
Results are shown in Fig. \ref{fig_olt_vs_rbt_2_8}. 
These were generated using four large publicly available datasets with over 100,000 frames in all to ensure their statistical significance.
The tracking performance is evaluated using success rate which is the fraction of total frames where the tracking error is less than a given threshold.
Tracking error is measured using alignment error \citep{Roy2015_tmt} that accounts for fine misalignments of pose better than conventional measures like Jaccard index or center location error.
More details about the datasets and the evaluation methodology can be found in \citep{singh17_ssim, singh17_mtf_thesis}.

As expected, all the OLTs have far lower SR than both 2 and 8 DOF RBTs for smaller $ t_p $ since they are less precise in general \citep{Kristan2016_vot16}. They do close the gap as $ t_p $ increases but none manage to surpass even the best 2 DOF RBTs, which themselves are notably worse than the 8 DOF ones.
This difference is even larger if OLTs and 2 DOF RBTs are evaluated against the full pose ground truth which is more indicative of the motions that need to be tracked in real tasks.
Though the superiority of DSST and Struck over other OLTs is consistent with VOT results \citep{Kristan2016_vot16}, the very poor performance of GOTURN \citep{Held2016_goturn}, which is one of the best trackers there, indicates a fundamental difference in the challenges involved within the two paradigms of tracking.
The speed plot shows another reason why OLTs are not suitable for high speed tracking applications - they are $ 10 $ to $ 30 $ times slower than the faster RBTs.


%% file: conclusion.tex
\section{Conclusions and Future Work}
\label{conclusions}
This paper presented MTF, a modular and extensible open source framework for RBT.
It provides highly efficient C++ implementations for several well established trackers that will hopefully address practical tracking needs of the wider robotics community. 
To this end, it has been designed to integrate well with popular libraries like ROS, OpenCV and ViSP so it can be easily used with existing as well as future projects that require fast trackers.
A novel method to decompose RBTs into sub modules was also formulated that facilitates the separat study of appearance models, state space, and search methods.
 

MTF is still a work in progress and offers several promising avenues of future extensions for each of the sub modules.
This includes novel composite SMs especially those that, like \texttt{GridTracker}, run a large number of relatively simple trackers simultaneously and combine their outputs to deduce the state of the tracked patch with greater precision and robustness than any single tracker can possibly provide.
A contribution in this direction is the \texttt{LineTracker} (Fig. \ref{fig_class_diagram}) \citep{singh17_mtf_thesis}. More generally a variety of multiple-view constraints, e.g. projective invariants, could be used to constrain and stabilize multiple individual MTF trackers.  

One of the most potentially beneficial ways to improve AMs is the incorporation of online learning to update the template.
As mentioned in Sec. \ref{sec_am}, MTF is designed to support this and two related AMs - \texttt{DFM} \citep{Siam15_dfm} and \texttt{PCA} \citep{Ross08ivt} - are already implemented that respectively utilize offline and online learning. However, more powerful learning methods should be implemented, for example utilizing deep neural networks.
ILMs can also be extended with better parameterization that can account for other sources of variations in the appearance of the object patch such as motion blur and occlusion.
Another promising extension is the ability to handle depth information from 3D cameras like Kinect whose increasing ubiquity may make this the easiest way to improve tracking performance.

SSMs can be improved by using motion learning from annotated sequences to generate better stochastic samples.
As mentioned in Sec. \ref{stochastic_search}, the performance of stochastic SMs depends largely on the quality of samples so any progress in this direction should definitely be worthwhile.
Further, addition of non rigid SSMs that can go beyond the basic planar projective transforms will be a useful extension albeit with somewhat limited application domain.
Similarly, SSMs that support 3D motion estimation are needed to complement the aforementioned depth information processing support in AMs.
Finally, improvements can be made to the implementations of existing methods to make them more practically useful. For instance, slower methods like \texttt{PF}, \texttt{MI}, \texttt{CCRE}, \texttt{NGF} and \texttt{GridTracker} can be efficiently parallelized.

%% file: appendix.tex

\appendix

\section{$ \hat{\mathbf{J}} $ and $ \hat{\mathbf{H}} $ for different variants of LK}
\label{appendix}
\subsection{Jacobian}
\label{jacobian}
Denoting $\mathbf{w}(\mathbf{x},\mathbf{p_s})$ with $ \mathbf{w}(\mathbf{p}) $ for conciseness ($ A=0 $ and $ \mathbf{x} $ is constant in this context) and letting  $\mathbf{\hat{p}_t}$ denote an estimate of $\mathbf{p_t}$ to which an incremental update is sought, the formulations for $ \hat{\mathbf{J}} $ used by FALK and FCLK are:
\begin{equation}
\label{eq_jac_falk}
 \hat{\mathbf{J}}_{fa} = 
 \left.\dfrac{\partial f}{\partial \mathbf{I^c}}\right\rvert_{\mathbf{I^c}=\mathbf{I_t}(\mathbf{w}(\mathbf{\hat{p}_t}))}
 \left.
  	 	\nabla\mathbf{I_t}
 \right\rvert_{\mathbf{x}=\mathbf{w}(\mathbf{\hat{p}_t})}  
 \left.
 	\dfrac{\partial \mathbf{w}}{\partial \mathbf{p}}
 \right\rvert_{\mathbf{p}=\mathbf{\hat{p}_t}}
\end{equation}
\begin{align}
\label{eq_jac_fclk}
 \hat{\mathbf{J}}_{fc} = 
 \left.\dfrac{\partial f}{\partial \mathbf{I^c}}\right\rvert_{\mathbf{I^c}=\mathbf{I_t}(\mathbf{w}(\mathbf{\hat{p}_t}))}
\left.
\nabla \mathbf{I_t} (\mathbf{w})
\right\rvert_{\mathbf{x}=\mathbf{x_0}}  
\left.
   \dfrac{\partial \mathbf{w}}{\partial \mathbf{p}}
\right\rvert_{\mathbf{p}=\mathbf{p_0}}
\end{align}
where $ \nabla \mathbf{I_t} (\mathbf{w}) $ in Eq. \ref{eq_jac_fclk} refers to the gradient of $ I_t $ warped using $\mathbf{\hat{p}_t}$,
i.e. $ I_t $ is first warped back to the coordinate frame of $ I_0 $ using $ \mathbf{w}(\mathbf{\hat{p}_t})$ to obtain $ I_t(\mathbf{w}) $ whose gradient is then computed at $ \mathbf{x}=\mathbf{x_0} $.
It can be further expanded \citep{Baker04lucasKanade_paper} as:
\begin{equation}
\label{eq_jac_warped_grad}
\left.
	\nabla \mathbf{I_t} (\mathbf{w})
\right\rvert_{\mathbf{x}=\mathbf{x_0}}  
=
 \left.
  	 	\nabla \mathbf{I_t}
 \right\rvert_{\mathbf{x}=\mathbf{w}(\mathbf{\hat{p}_t})} 
\left.
	\dfrac{\partial \mathbf{w}}{\partial \mathbf{x}}
\right\rvert_{\mathbf{p}=\mathbf{\hat{p}_t}}
\end{equation}
Since $ \nabla \mathbf{I_t} $ is usually the most computationally intensive part of $  \mathbf{J}_{fc} $ and $  \mathbf{J}_{fa} $, the so called inverse methods approximate this with the gradient of $ \nabla \mathbf{I_0} $ for efficiency as this only needs to be computed once. The specific expressions for these methods are:
\begin{equation}
\label{eq_jac_iclk}
 \hat{\mathbf{J}}_{ic} =
 \left.
 	\dfrac{\partial f}{\partial \mathbf{I^*}}
 \right\rvert_{\mathbf{I^*}=\mathbf{I_0}(\mathbf{x_0})}
 \left.
 	\nabla \mathbf{I_0}
 \right\rvert_{\mathbf{x}=\mathbf{x_0}}  
 \left.
 	\dfrac{\partial \mathbf{w}}{\partial \mathbf{p}}
 \right\rvert_{\mathbf{p}=\mathbf{p_0}}
\end{equation}
\begin{equation}
\label{eq_jac_ialk}
 \hat{\mathbf{J}}_{ia} = 
 \left.\dfrac{\partial f}{\partial \mathbf{I^c}}\right\rvert_{\mathbf{I^c}=\mathbf{I_t}(\mathbf{w}(\mathbf{\hat{p}_t}))}
 \left.\nabla \mathbf{I_0}\right\rvert_{\mathbf{x}=\mathbf{x_0}}
 \left.\dfrac{\partial \mathbf{w}}{\partial \mathbf{x}}^{-1}\right\rvert_{\mathbf{p}=\mathbf{\hat{p}_t}}  
 \left.\dfrac{\partial \mathbf{w}}{\partial \mathbf{p}}\right\rvert_{\mathbf{p}=\mathbf{\hat{p}_t}}
\end{equation}
where the middle two terms in Eq. \ref{eq_jac_ialk} are derived from Eqs. \ref{eq_jac_falk} and \ref{eq_jac_warped_grad} by assuming \citep{Hager98parametricModels} that $\mathbf{w}(\mathbf{\hat{p}_t})$ perfectly aligns $ I_t $ with $ I_0 $, i.e.
$ I_t(\mathbf{w}) = I_0 $ so that
\begin{align}
\label{eq_ialk_assumption}
\nabla \mathbf{I_t} (\mathbf{w})=\nabla \mathbf{I_0}
\end{align}
In its original paper \citep{Benhimane07_esm_journal}, ESM was formulated as using the mean of the pixel gradients $\nabla \mathbf{I_0}$ and $\nabla \mathbf{I_t}(\mathbf{w})$ to compute $ \mathbf{J} $ but, as this formulation is only applicable to SSD, we consider a generalized version \citep{Brooks10_esm_ic, Scandaroli2012_ncc_tracking} that uses the \textit{difference} between FCLK and ICLK Jacobians:
\begin{align}
\label{eq_jac_esm}
 \hat{\mathbf{J}}_{esm} =  \hat{\mathbf{J}}_{fc} -   \hat{\mathbf{J}}_{ic}
\end{align}

\subsection{Hessian}
\label{hessian}

For clarity and brevity, evaluation points for the various terms have been omitted in the equations that follow as being obvious from analogy with the previous section.

It is generally assumed \citep{Baker04lucasKanade_paper, Benhimane07_esm_journal} that the second term of Eq. \ref{eq_hess_basic} is too costly to compute and too small near convergence to matter and so is omitted to give the so called Gauss Newton Hessian
\begin{align}
\label{eq_hess_basic_gn}
\mathbf{H}_{gn}
=
\dfrac{\partial \mathbf{I}}{\partial \mathbf{p}}^T
\dfrac{\partial^2 f}{\partial \mathbf{I}^2}
\dfrac{\partial \mathbf{I}}{\partial \mathbf{p}}
\end{align}
Though $ \mathbf{H}_{gn} $ works very well for SSD (and in fact even better than $ \mathbf{H}$ \citep{Baker04lucasKanade_paper,Dame10_mi_ict}), it is well known \citep{Dame10_mi_ict, Scandaroli2012_ncc_tracking} to \textit{not} work well with other AMs like MI, CCRE and NCC for which an approximation to the Hessian \textit{after convergence} has to be used by assuming perfect alignment or $ \mathbf{I_t}(\mathbf{w}(\mathbf{\hat{p}_t})) = \mathbf{I_0}(\mathbf{x_0})$. We refer to the resultant approximation as the \textbf{Self Hessian} $ \mathbf{H}_{self} $ and, as this substitution can be made by setting either $ \mathbf{I^c} = \mathbf{I_0}(\mathbf{x_0}) $ or $ \mathbf{I^*} = \mathbf{I_t}(\mathbf{w}(\mathbf{\hat{p}_t})) $, we get two forms which are respectively deemed to be the Hessians for ICLK and FCLK:
\begin{align}
\label{eq_hess_iclk}
\hat{\mathbf{H}}_{ic}
=
\mathbf{H}_{self}^\mathbf{*}
=
\dfrac{\partial \mathbf{\mathbf{I_0}}}{\partial \mathbf{p}}^T
\dfrac{\partial^2 f(\mathbf{I_0}, \mathbf{I_0})}{\partial \mathbf{I}^2}
\dfrac{\partial \mathbf{\mathbf{I_0}}}{\partial \mathbf{p}}
+
\dfrac{\partial f(\mathbf{I_0}, \mathbf{I_0})}{\partial \mathbf{I}}
\dfrac{\partial^2 \mathbf{\mathbf{I_0}}}{\partial \mathbf{p}^2}
\end{align}
\begin{align}
\label{eq_hess_fclk}
\hat{\mathbf{H}}_{fc}
=
\mathbf{H}_{self}^\mathbf{c}
=
\dfrac{\partial \mathbf{\mathbf{I_t}}}{\partial \mathbf{p}}^T
\dfrac{\partial^2 f(\mathbf{I_t}, \mathbf{I_t})}{\partial \mathbf{I}^2}
\dfrac{\partial \mathbf{\mathbf{I_t}}}{\partial \mathbf{p}}
+
\dfrac{\partial f(\mathbf{I_t}, \mathbf{I_t})}{\partial \mathbf{I}}
\dfrac{\partial^2 \mathbf{\mathbf{I_t}}}{\partial \mathbf{p}^2}
\end{align}
It is interesting to note that $ \mathbf{H}_{gn} $ has the exact same form as $ \mathbf{H}_{self} $ for SSD (since $ \dfrac{\partial f_{ssd}(\mathbf{I_0}, \mathbf{I_0})}{\partial \mathbf{I}} =  \dfrac{\partial f_{ssd}(\mathbf{I_t}, \mathbf{I_t})}{\partial \mathbf{I}} = \mathbf{0}$) so it seems that interpreting Eq. \ref{eq_hess_basic_gn} as the first order approximation of Eq. \ref{eq_hess_basic}, as in \citep{Baker04lucasKanade_paper, Dame10_mi_ict}, is incorrect and it should instead be seen as a special case of $ \mathbf{H}_{self} $.

$ \hat{\mathbf{H}}_{fa} $ differs from $ \hat{\mathbf{H}}_{fc} $ only in the way $ \dfrac{\partial^2 \mathbf{\mathbf{I_t}}}{\partial \mathbf{p}^2} $ and $ \dfrac{\partial \mathbf{\mathbf{I_t}}}{\partial \mathbf{p}} $ are computed for the two as given in Eqs. \ref{eq_pix_hess_falk} and \ref{eq_pix_hess_fclk} respectively.
\begin{align}
\label{eq_pix_hess_falk}
\dfrac{\partial^2 \mathbf{\mathbf{I_t}}}{\partial \mathbf{p}^2} (fa)
= 
\dfrac{\partial\mathbf{w}}{\partial \mathbf{p}}^T
\nabla^2 \mathbf{I_t}
\dfrac{\partial \mathbf{w}}{\partial \mathbf{p}}
+
\nabla\mathbf{I_t}
\dfrac{\partial^2 \mathbf{w}}{\partial \mathbf{p}^2}
\end{align}
\begin{align}
\label{eq_pix_hess_fclk}
\dfrac{\partial^2 \mathbf{\mathbf{I_t}}}{\partial \mathbf{p}^2} (fc)
= 
\dfrac{\partial\mathbf{w}}{\partial \mathbf{p}}^T
\nabla^2 \mathbf{I_t}(\mathbf{w})
\dfrac{\partial \mathbf{w}}{\partial \mathbf{p}}
+
\nabla\mathbf{I_t}(\mathbf{w})
\dfrac{\partial^2 \mathbf{w}}{\partial \mathbf{p}^2}
\end{align}
where $\nabla^2 \mathbf{I_t}(\mathbf{w})$ can be expanded by differentiating Eq. \ref{eq_jac_warped_grad} as:
\begin{align}
\label{eq_warped_pix_hess}
\nabla^2 \mathbf{I_t}(\mathbf{w})
=
\dfrac{\partial \mathbf{w}}{\partial \mathbf{x}}^T
\nabla^2 \mathbf{I_t}
\dfrac{\partial \mathbf{w}}{\partial \mathbf{x}}
+
\nabla\mathbf{I_t}
\dfrac{\partial^2 \mathbf{w}}{\partial \mathbf{x}^2}
\end{align}
$ \hat{\mathbf{H}}_{ia} $ is identical to $ \hat{\mathbf{H}}_{fa} $ except that $ \nabla\mathbf{I_0} $ and $ \nabla^2\mathbf{I_0} $ are used to approximate $ \nabla\mathbf{I_t} $ and $ \nabla^2\mathbf{I_t}$. The expression for the former is in Eq. \ref{eq_jac_ialk} while that for the latter can be derived by differentiating both sides of Eq. \ref{eq_jac_warped_grad} after substituting Eq. \ref{eq_ialk_assumption}:
\begin{align*}
\nabla^2 \mathbf{I_0}
=
\dfrac{\partial \mathbf{w}}{\partial \mathbf{x}}^T
\nabla^2 \mathbf{I_t}
\dfrac{\partial \mathbf{w}}{\partial \mathbf{x}}
+
\nabla\mathbf{I_t}
\dfrac{\partial^2 \mathbf{w}}{\partial \mathbf{x}^2}
\end{align*}
which gives:
\begin{align}
\label{eq_ialk_pix_hess}
\nabla^2 \mathbf{I_t}(ia)
&=
\left(\dfrac{\partial \mathbf{w}}{\partial \mathbf{x}}^{-1}\right)^T
\left[
\nabla^2 \mathbf{I_0}
-
\nabla\mathbf{I_t}
\dfrac{\partial^2 \mathbf{w}}{\partial \mathbf{x}^2}
\right]
\dfrac{\partial \mathbf{w}}{\partial \mathbf{x}}^{-1}
\nonumber\\
&=
\left(\dfrac{\partial \mathbf{w}}{\partial \mathbf{x}}^{-1}\right)^T
\left[
\nabla^2 \mathbf{I_0}
-
\left(
\nabla\mathbf{I_0}
\dfrac{\partial \mathbf{w}}{\partial \mathbf{x}}^{-1}
\right)
\dfrac{\partial^2 \mathbf{w}}{\partial \mathbf{x}^2}
\right]
\dfrac{\partial \mathbf{w}}{\partial \mathbf{x}}^{-1}
\end{align}
where the second equality again follows from Eq. \ref{eq_jac_warped_grad} and \ref{eq_ialk_assumption}.
Finally, the ESM Hessian corresponding to the Jacobian in Eq. \ref{eq_jac_esm} is the \textit{sum} of FCLK and ICLK Hessians:
\begin{align}
\label{eq_hess_esm}
\hat{\mathbf{H}}_{esm} = \hat{\mathbf{H}}_{fc} +  \hat{\mathbf{H}}_{ic}
\end{align}
